\documentclass[journal]{IEEEtran}  
\usepackage{enumerate}
\usepackage{url}
\usepackage{flushend}
\usepackage{multirow}
\usepackage{colortbl}
\usepackage{pifont}
\usepackage{setspace}
\usepackage{cite}
\usepackage{algorithm}
\usepackage{algorithmic}
\usepackage[subfigure]{graphfig}
\usepackage{amsthm,amsmath,amssymb}
\usepackage{subfigure}
\usepackage{graphicx}
\usepackage{mathrsfs}
\usepackage{makecell}
\usepackage{cleveref}
\usepackage{booktabs}

\begin{document}
\graphicspath{{simulationresults/}}

\title{Self-learning sparse PCA for multimode process monitoring}

\author{Jingxin Zhang, Donghua Zhou, ~\IEEEmembership{Fellow,~IEEE}, and Maoyin Chen, ~\IEEEmembership{Member,~IEEE}
	\thanks{This work was supported by National Natural Science Foundation of China [grant numbers 62033008, 61873143]. (Corresponding authors: Donghua Zhou; Maoyin Chen)}
	\thanks{Jingxin Zhang is with the Department of Automation, Tsinghua University, Beijing 100084, China (e-mail: zjx18@mails.tsinghua.edu.cn). }
	\thanks{Donghua Zhou is with College of Electrical Engineering and Automation, Shandong University of Science and Technology, Qingdao 266000, China and also with the Department of Automation, Tsinghua University, Beijing 100084, China (e-mail: zdh@mail.tsinghua.edu.cn).}
	\thanks{Maoyin Chen is with the Department of Automation, Tsinghua University, Beijing 100001, China and also with School of Automation and Electrical Engineering, Linyi University, Linyi 276005, China (e-mail: mychen@tsinghua.edu.cn).
	}
\thanks{This paper has been submitted for IEEE Transactions on Industrial Informatics for potential publication.}
}

\maketitle
\IEEEpeerreviewmaketitle

\begin{abstract}	
	This paper proposes a novel sparse principal component analysis  algorithm with self-learning ability for successive modes, where synaptic intelligence is employed to measure the importance of variables and a regularization term is added to preserve the learned knowledge of previous modes. Different from traditional multimode monitoring  methods, the monitoring model is updated based on the current model and new data when a new mode arrives,  thus delivering prominent performance for sequential modes. Besides, the computation and storage resources are saved in the long run, because it is not necessary to retrain the model from scratch frequently and store data from previous modes. More importantly, the model furnishes excellent interpretability owing to the sparsity of parameters. 	Finally, a numerical case and a practical pulverizing system are adopted to illustrate the effectiveness of the proposed algorithm.
	
\end{abstract}


\section{Background}
Multimode process monitoring is increasingly demanded and significant, as industrial systems generally operate in varying modes due to raw materials, changing load, etc \cite{quiones-grueiro2019data,wang2020data,zhang2018a}. Data from different modes have different characteristics, such as mean value and variance \cite{zhou2018multimode,liu2020a}. It is imperative to investigate the effective manners for multimode processes \cite{jiang2019recent,zhang2019process}.

Marcos \emph{et al} \cite{quiones-grueiro2019data} summarized the techniques for multimode processes and divided the methods into two major categories, namely, single-model schemes and multiple-model schemes. Single-model methods aim to find a transformation to remove the multimodality features and then the fault is detected by a decision function \cite{zhang2018a}.  Multiple-model methods  identify the mode and build the monitoring model within each mode \cite{peng2017multimode,wang2020data,shang2019incipient}. However, the information on all modes should be complete, which is evidently impossible in real systems.
When a new mode arrives, the learned knowledge of previous modes may be overwritten when training  the same monitoring model for the current mode, which would lead to a abrupt performance decrease. This phenomenon has been  analyzed and discussed in \cite{zhang2020multimode}. Overall, state-of-the-art methods aforementioned need to retrain the monitoring model from scratch \cite{zhang2020multimode,KirkpatrickOvercoming}.

Since the operating modes in practical systems appear successively, it is consuming to progressively retain massive data  and repeatedly retrain the models.  Therefore, it is important to establish a self-learning monitoring model and update it continually when a new mode arrives.  One strategy  of realizing self-learning ability is continual learning, where the model is updated when new data are available. The technical core of continual learning is to accommodate new information while preserving the  acquired knowledge  \cite{aljundi2019task,KirkpatrickOvercoming}. However, there exists one longstanding challenge, namely, `catastrophic forgetting' issue, where the information of previous modes is overlapped by new data and a new model based on new data may fail to monitor previous modes. Various schemes have been developed to alleviate this issue and supply excellent performance \cite{aljundi2019task,KirkpatrickOvercoming,Raia2020Embracing,Jaehong2018Lifelong}.   Nevertheless, it is still scarce to investigate the process monitoring techniques with self-learning ability \cite{xu2017self,ye2015self,feldmann2019all}.  %

 Zhang \emph{et al} firstly focused on this research and illustrated the necessity in \cite{zhang2020multimode}, where  a single monitoring model with continual learning ability was investigated for successive modes.  Elastic weight consolidation (EWC) \cite{KirkpatrickOvercoming} was employed to solve the `catastrophic forgetting' of traditional principal component analysis (PCA) \cite{zhang2020multimode,zhang2021Monitoring}, referred to as PCA-EWC,  which retained the significant information of previous modes by slowing down the changes of certain influential parameters. Note that EWC estimates the importance measure offline based on the point estimate of Fisher information matrix (FIM) \cite{KirkpatrickOvercoming}.

 \begin{figure*}[!tbp]   
 	\centering
 	\subfigure{\label{fig1}}\addtocounter{subfigure}{-2}
 	\subfigure
 	{\subfigure[]{\includegraphics[width=0.26\textwidth]{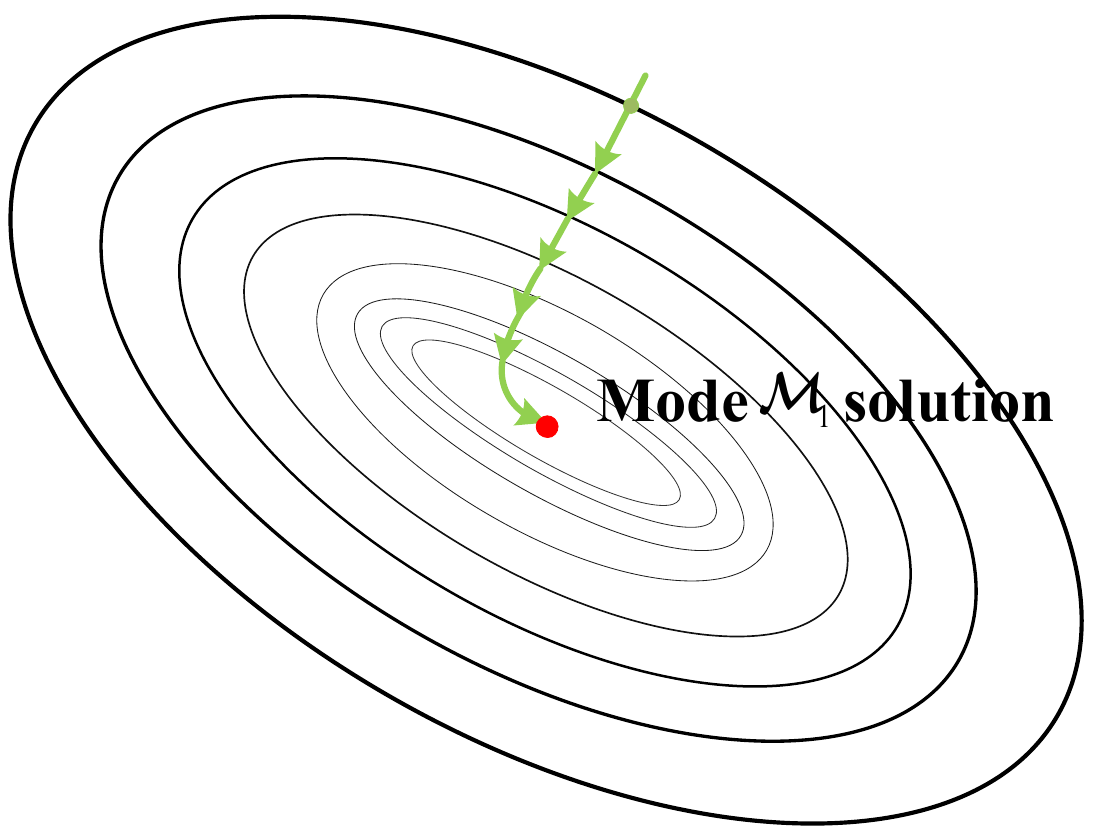}}}
 	\hspace{-1mm}
 	\subfigure{\label{fig2}}\addtocounter{subfigure}{-2}
 	\subfigure
 	{\subfigure[]{\includegraphics[width=0.33\textwidth]{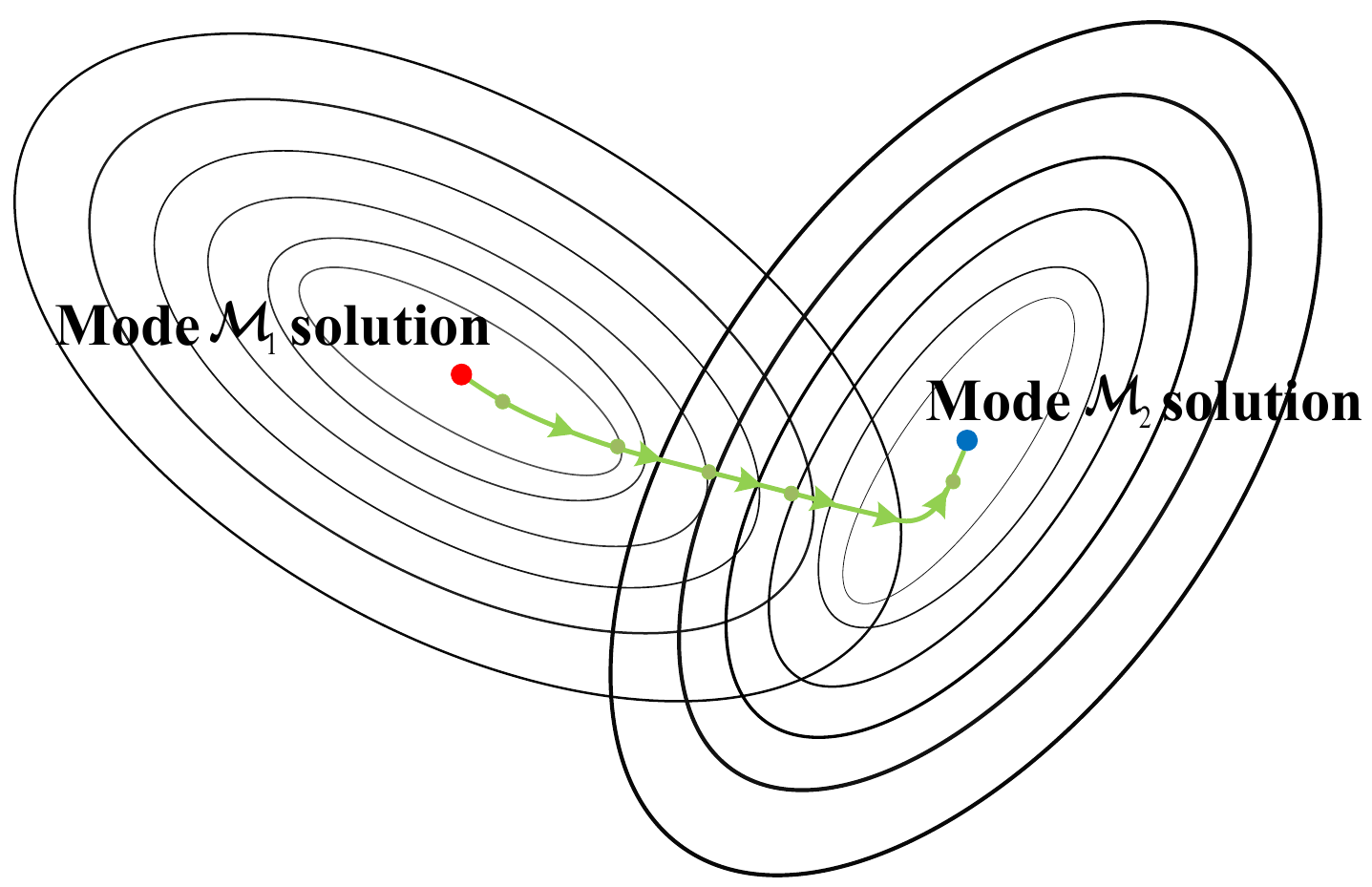}}}
 	\subfigure{\label{fig3}}\addtocounter{subfigure}{-2}
 	\subfigure
 	{\subfigure[]{\includegraphics[width=0.31\textwidth,]{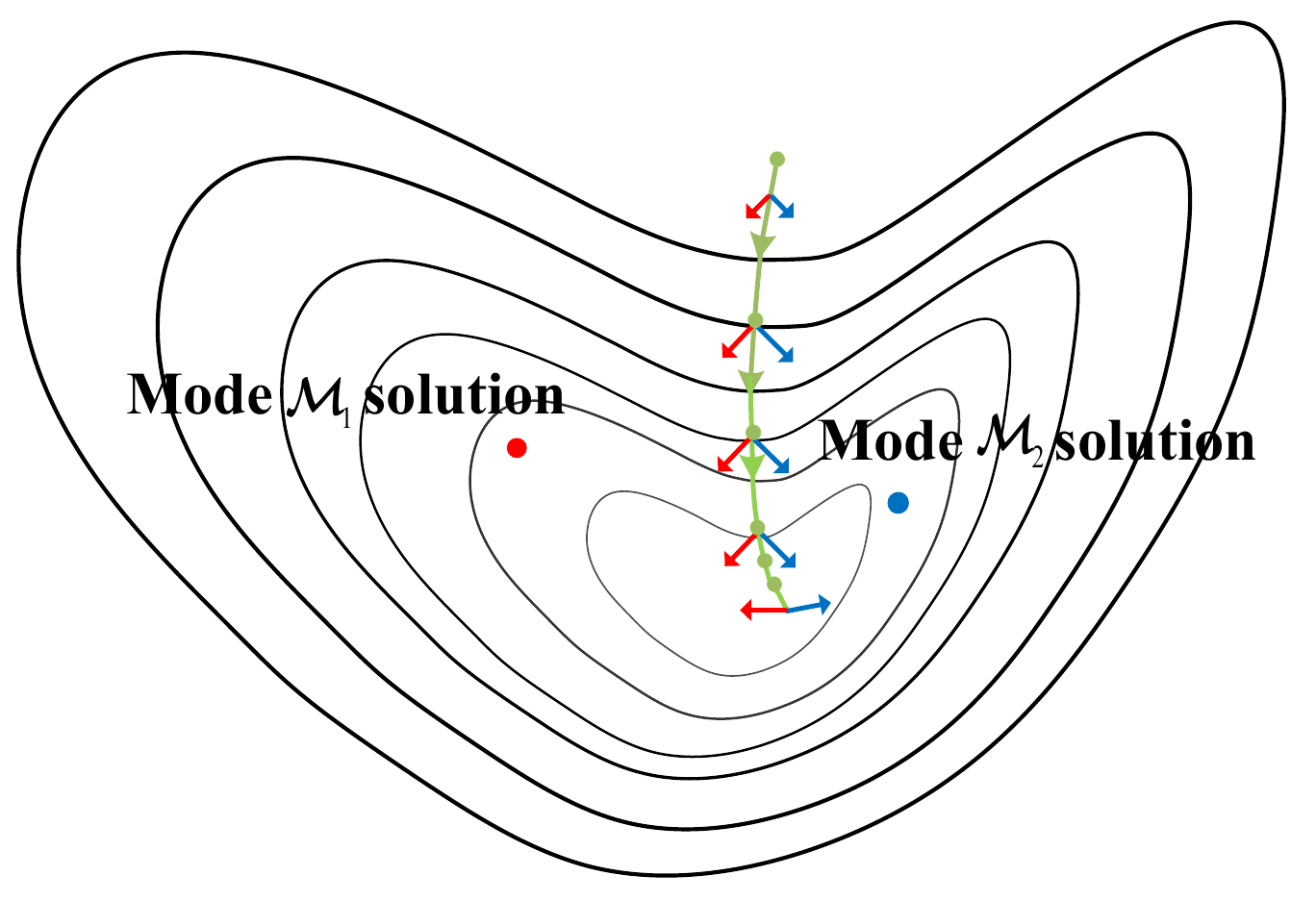}}}
 	\centering
 	\caption{Illustration of gradient decent optimization for different modes: (a) The trajectory for one mode; (b) The trajectory when training the same model on the second mode subsequently; (c) The trajectory when minimizing the total loss from both modes (green) and  gradients from each mode (red and blue) \cite{Raia2020Embracing}.} \label{fig_gradient}
 \end{figure*}

Consider the poor interpretability of PCA, this paper investigates sparse PCA (SPCA) with self-learning ability for multimode process monitoring, where the model is updated based on the current model and new data.
%
Instead of EWC, we consider synaptic intelligence (SI), which calculates the importance matrix along the entire learning trajectory \cite{zenke2017continual,MasseE10467Alleviating,ven2020brain}, by computing the gradients of loss and the parameter update. For convenience, the proposed SPCA with SI is denoted as SPCA-SI. 
%
Compared with PCA-EWC \cite{zhang2020multimode},  SPCA-SI utilizes $L_1$ regularization to enhance the sparsity of critical parameters, thus  providing better model interpretability. Besides, the importance measure by SI is easier to estimate than EWC, because the gradients are usually available while  FIM is intractable. Moreover, sparse representation is also beneficial to reduce catastrophic forgetting, as there are fewer model-sensitive parameters \cite{Raia2020Embracing,MasseE10467Alleviating,Jaehong2018Lifelong}.
In this paper, we assume that the mode transition is accomplished in a short time  and the switching time is available.

The rest of this paper is organized below. Section \ref{sec:2} reviews SPCA and SI briefly, and introduces the research problem. The proposed SPCA-SI algorithm is elaborated in Section \ref{sec:3} and settled by accelerated proximal gradient descent (APG) method. A novel $T^2$ statistic is proposed and the monitoring procedure is summarized in Section \ref{sec:statistic}. Besides, the influence of parameters and computational complexity are discussed.
The effectiveness of the proposed approach is illustrated by a numerical case study and a practical coal puzzling system in Section \ref{sec:4}. The concluding remark is given in Section \ref{sec:5}.

 \section{Preliminary}\label{sec:2}

\subsection{Revisit of SPCA}

Here we give another perspective of SPCA, where the projection vectors are acquired one by one.

Given the dataset $\boldsymbol  X \in R^{N \times m}$, $N$ is the number of samples and $m$ is the number of variables. PCA aims to minimize the reconstruction error, namely, 
\begin{center}
 $ \begin{array}{l}
\min \;\left\| {\boldsymbol X - \boldsymbol X \boldsymbol p{\boldsymbol p^T}} \right\|_F^2\\
s.t.\quad {\boldsymbol p^T}\boldsymbol p = 1
\end{array}$
\end{center}
where $\boldsymbol p \in R^{m}$ is the projection vector.

Consider the virtues of sparsity, $L_1$ regularization is adopted to enhance interpretability and alleviate catastrophic forgetting simultaneously \cite{Raia2020Embracing,Jaehong2018Lifelong}. Thus, the objective of SPCA is
\begin{equation}\label{originalaim_v3}
  \begin{array}{l}
\min \;\left\| {\boldsymbol X - \boldsymbol X \boldsymbol p{\boldsymbol p^T}} \right\|_F^2+\lambda \left\|\boldsymbol p\right\|_1 \\
s.t.\quad {\boldsymbol p^T}\boldsymbol p = 1
\end{array}
\end{equation}
where $\lambda$ is  a regularization parameter. After $\boldsymbol p$ is calculated, 
let $\boldsymbol X = \boldsymbol X - \boldsymbol X \boldsymbol p{\boldsymbol p^T}$, and repeat (\ref{originalaim_v3}) until $l$ projection vectors are obtained. Here, the number of principal components  $l$ is determined by cumulative percent variance. 

%

\subsection{Review of SI}

The synaptic framework was proposed  and detailed information has been described in \cite{zenke2017continual}. Here we overview the key points \cite{zenke2017continual,MasseE10467Alleviating}. 

For a learning process, we aim to seek for the optimal parameter $\boldsymbol \theta$ given the objective function $J(\boldsymbol \theta)$. Gradient-based methods are effective manners to solve the optimization problem. SI estimates the importance measure for each parameter along the learning trajectory, which reflects the sensitivity of each parameter to the loss.

The gradient is a conservative field, and the value of the integral along the trajectory is equal to the difference between the end point and the start point \cite{zenke2017continual}. 
 Consider an infinitesimal parameter update $\boldsymbol \delta \left( k \right)$ at $k$th iteration, the change in loss is approximated by
\begin{center}
 $ J\left( {\boldsymbol \theta \left( k \right) + \boldsymbol \delta \left( k \right)} \right) - J\left( {\boldsymbol \theta \left( k \right)} \right) \approx \sum\nolimits_i {{g_i}\left( k \right){\delta _i}\left( k \right)}$
\end{center}
where $\boldsymbol g = \frac{{\partial J}}{{\partial {\boldsymbol \theta }}}$ is the gradient, 
${\delta _i}\left( k \right) = \theta _i\left( k \right)-\theta _i\left( k-1 \right)$. 
During the whole learning process, the change in loss over the entire trajectory is calculated by
\begin{center}
$ \begin{aligned}
\sum\limits_k {\boldsymbol g\left( k \right) \boldsymbol \delta \left( k \right)}  =& \sum\limits_i {\sum\limits_k {{g_i}\left( k \right){\delta _i}\left( k \right)} } \\
 = & - \sum\limits_i {{\varpi _i}}
\end{aligned}$
\end{center}
More intuitively \cite{ven2020brain},
\begin{equation}\label{importance_mearure_v1}
  {\varpi _i} = \sum\limits_k {\left( {{\theta _i}\left( k \right) - {\theta _i}\left( {k - 1} \right)} \right)\frac{{ - \partial J}}{{\partial {\theta _i\left( k \right)}}}}
\end{equation}
Then, the importance measure is normalized by \cite{MasseE10467Alleviating}
\begin{equation}
{{\bar \varpi }_i} = \max \left( {0,\frac{{{\varpi _i}}}{{{{\left( {\Delta {\theta _i}} \right)}^2} + \zeta }}} \right)
\end{equation}
thus the regularization term and the loss shares the same unit.
 $\Delta {\theta _i}{\rm{ = }}\sum\nolimits_k {\left( {{\theta _i}\left( k \right) - {\theta _i}\left( {k - 1} \right)} \right)} $ is the total change for each parameter and $ \zeta $ is added to avoid ill-conditioning issue.


\subsection{Problem reformulation}
This paper proposes a self-learning process monitoring approach  for successive modes, which is built based on SPCA and the model is updated when a new mode arrives.
We take two modes to depict the research problem by Fig. \ref{fig_gradient} \cite{Raia2020Embracing}. 

When training the monitoring model for mode $\mathcal{M}_1$, the optimization issue is settled by gradient decent method and the trajectory of SPCA is exhibited in Fig. \ref{fig1}. When a new mode $\mathcal{M}_2$ arrives, traditional SPCA-based methods generally train the model subsequently, as illustrated in  Fig. \ref{fig2}. In this case, the learning of mode $\mathcal{M}_2$ leads to an overlap of the learned knowledge, which indicates that  the retrained model is not efficient for the previous mode $\mathcal{M}_1$. This paper aims to accommodate new data by a continually updated model  while accumulating the learned knowledge, 
thus delivering brilliant performance for  two or  more modes.
As shown in Fig. \ref{fig3}, the total loss for both modes is considered simultaneously and the optimal solution is an equilibrium between the gradients of different modes \cite{Raia2020Embracing}.

\section{Methodology}\label{sec:3}

In this section, we present the procedure of SPCA-SI for the first mode and the objective is optimized by APG. Then, SPCA-SI is extended to more general cases.

\subsection{SPCA-SI for the first mode}
For the first mode $\mathcal{M}_1$,  data $\boldsymbol X_1$ are collected. We settle the issue (\ref{originalaim_v3}) by augmented Lagrangian function:
%
\begin{equation}\label{LAGRANGE}
  J = \left\| {\boldsymbol X_1 - \boldsymbol X_1 \boldsymbol p{\boldsymbol p^T}} \right\|_F^2 + \lambda {\left\|\boldsymbol p \right\|_1} + \mu {\left( {\boldsymbol p^T}\boldsymbol p - 1\right)^2}
\end{equation}
where $\mu $ is the Lagrange parameter. (\ref{LAGRANGE}) is nonconvex and nonsmooth, and can not be directly settled by gradient-based methods.

  APG is an effective technique to deal with this type of optimization issue and employed in this paper \cite{li2015accelerated}.
 We divide (\ref{LAGRANGE}) into smooth part $g\left( \boldsymbol p \right)$ and nonsmooth part $h\left( \boldsymbol p \right)$, namely,
\begin{equation}\label{smoothg}
  g\left( \boldsymbol p \right)=\left\| {\boldsymbol X_1 - \boldsymbol X_1 \boldsymbol p \boldsymbol p^T} \right\|_F^2 +\mu {\left( {\boldsymbol p^T}\boldsymbol p - 1\right)^2}
\end{equation}
\begin{equation}\label{nonsmoothh}
h\left( \boldsymbol p \right)=\lambda {\left\| \boldsymbol p \right\|_1}
\end{equation}

\subsection{Solution with APG}
The procedure of APG contains the gradient-based part of $g\left( \boldsymbol p \right)$ and the proximal gradient. 



For the smooth part $g\left( \boldsymbol p \right)$, (\ref{smoothg}) is equivalent to
\begin{center}
$\begin{aligned}
 g\left( \boldsymbol p \right) = &tr\left( {\boldsymbol X_1^T}\boldsymbol X_1 \right) + \mu  + tr\left( {\boldsymbol p{\boldsymbol p^T}\left( {{\boldsymbol X_1^T}\boldsymbol X_1 + \mu \boldsymbol I} \right)\boldsymbol p{\boldsymbol p^T}} \right) \\
 &- 2tr\left( {\boldsymbol p{\boldsymbol p^T}\left( {{\boldsymbol X_1^T}\boldsymbol X_1 + \mu \boldsymbol  I} \right)} \right)
 \end{aligned}$
\end{center}
Thus, the gradients are
\begin{equation} \label{gradp1}
 {\nabla_{\boldsymbol p}}g\left( \boldsymbol p \right) = \frac{{\partial g}}{{\partial {\boldsymbol p}}} = \boldsymbol p{\boldsymbol p^T} \boldsymbol G_1 \boldsymbol p + \boldsymbol G_1 \boldsymbol p{\boldsymbol p^T}\boldsymbol p - 2\boldsymbol G_1 \boldsymbol p
\end{equation}
\begin{equation}\label{gradmu}
  {\nabla_{\mu} }g\left( \boldsymbol p \right)=\frac{{\partial g}}{{\partial \mu }} = \left( {\boldsymbol p^T}\boldsymbol p - 1\right)^2
\end{equation}
where $\boldsymbol G_1=2\left({{\boldsymbol X_1^T}\boldsymbol X_1 + \mu \boldsymbol  I}\right)$.

For  $h\left( \boldsymbol p \right)$, the proximal function is defined as \cite{li2015accelerated,beck2009fast}
\begin{equation}\label{proximal1}
	\begin{aligned}
		{\boldsymbol p ^ + } = &\arg \mathop {\min }\limits_{\boldsymbol z} \frac{1}{{2t}}\left\| {\boldsymbol z - \left( {\boldsymbol p  - t\nabla_{\boldsymbol p} g\left( \boldsymbol p  \right)} \right)} \right\|_2^2 + h\left(\boldsymbol z \right)\\
		: =& pro{\boldsymbol x_{h,t}}\left( {\boldsymbol p  - t\nabla_{\boldsymbol p} g\left(\boldsymbol p  \right)} \right)
	\end{aligned}
\end{equation}
Inspired by Adam \cite{kingma2015adam},
the learning rate $t$  is adaptively estimated to accelerate convergence. At $k$th iteration, $t_k$ is calculated by
\begin{equation}\label{learningrate}
	\begin{aligned}
		{t_k} =& f\left( {\alpha ,{t_{k - 1}},\nabla {g_k}} \right)\\
		=& {\alpha  \mathord{\left/
				{\vphantom {\alpha  {\left( {\sqrt {{{\left( {{\tau _2}{t_{k - 1}} + \left( {1 - {\tau _1}} \right){{\left\| {\nabla {g_k}} \right\|}^2}} \right)} \mathord{\left/
											{\vphantom {{\left( {{\tau _2}{t_{k - 1}} + \left( {1 - {\tau _1}} \right){{\left\| {\nabla {g_k}} \right\|}^2}} \right)} {\left( {1 - {\tau _2}} \right)}}} \right.
											\kern-\nulldelimiterspace} {\left( {1 - {\tau _2}} \right)}}}  + \varepsilon } \right)}}} \right.
				\kern-\nulldelimiterspace} {\left( {\sqrt {{{\left( {{\tau _2}{t_{k - 1}} + \left( {1 - {\tau _1}} \right){{\left\| {\nabla {g_k}} \right\|}^2}} \right)} \mathord{\left/
								{\vphantom {{\left( {{\tau _2}{t_{k - 1}} + \left( {1 - {\tau _1}} \right){{\left\| {\nabla {g_k}} \right\|}^2}} \right)} {\left( {1 - {\tau _2}} \right)}}} \right.
								\kern-\nulldelimiterspace} {\left( {1 - {\tau _2}} \right)}}}  + \varepsilon } \right)}}
	\end{aligned}
\end{equation}
where $\nabla {g_k}$ is the corresponding gradient.
$\alpha$, ${\tau _1}$ and ${\tau _2}$ are constants.  $\varepsilon$ is added to avoid ill-conditioning issue.

\begin{algorithm}[!tbp]
	\caption{APG for optimization issue (\ref{LAGRANGE})}\label{APG}
	\textbf{Input:} Initialize $ \boldsymbol p_1 = \boldsymbol p_0$, $\boldsymbol z_1 = \boldsymbol p_0$, $t_1=t_0=0$, ${\boldsymbol {\bar \omega}}_1 = \boldsymbol 0$, $\tau_1 = 0.9$, $\tau_2 = 0.999$, $\varepsilon = 10^{-8}$, $t^y_{0} = 10^{-4}$, $t^p_{0} = 10^{-4}$, $t^{\mu}_0 = 10^{-4}$, $\alpha^p = 0.001$, $\alpha^{\mu} = 0.01$  \\
	\textbf{Output:} the optimal $\boldsymbol p$, and the importance measure $\boldsymbol \omega$\\
	\textbf{for} $k=1,2,3,\cdots$ \textbf{do}
\begin{enumerate}
		\item Update the projection vector:
		\begin{enumerate}  		
			\item  ${\boldsymbol y_k} = {\boldsymbol p_k} + \frac{{{t_{k - 1}}}}{{{t_k}}}\left( {{\boldsymbol z_k} - {\boldsymbol p_k}} \right) + \frac{{{t_{k - 1}} - 1}}{{{t_k}}}\left( {{\boldsymbol p_k} - {\boldsymbol p_{k - 1}}} \right)$
			\item ${\boldsymbol z_{k + 1}} = pro{x_{h,{t^y}}}\left( {{\boldsymbol y_k} - {t^y_{k}}\nabla_{\boldsymbol p} g\left( {{\boldsymbol y_k}} \right)} \right)$,
			calculate $t_k^y = f\left( {{\alpha^p},t_{k - 1}^y,{\nabla _{\boldsymbol p}}g\left( {{\boldsymbol y_k}} \right)} \right)$ by (\ref{learningrate})
			\item  ${\boldsymbol v_{k + 1}} = pro{x_{h,{t^p}}}\left( {{\boldsymbol p_k} - {t^p_k}\nabla_{\boldsymbol p} g\left( {{\boldsymbol p_k}} \right)} \right)$,
			calculate $t_k^p = f\left( {{\alpha ^p},t_{k - 1}^p,{\nabla _{\boldsymbol p}}g\left( {{\boldsymbol p_k}} \right)} \right)$ by (\ref{learningrate})
			\item $ {t_{k + 1}} = \frac{{\sqrt {4{{\left( {{t_k}} \right)}^2} + 1}  + 1}}{2}$
			\item  ${\boldsymbol p_{k + 1}} = \left\{ {\begin{array}{*{20}{c}}
					{{\boldsymbol z_{k + 1}},}&{if\;J\left( {{\boldsymbol z_{k + 1}}} \right) \le J\left( {{\boldsymbol v_{k + 1}}} \right)}\\
					{{\boldsymbol v_{k + 1}},}&{otherwise}
			\end{array}} \right.$
		\end{enumerate}
		\item Update $\mu$, $\mu_{k+1}= \mu_k + t^{\mu}_k{\nabla _{\mu}}g\left( \boldsymbol p_{k+1} \right)$,
		$t_k^\mu  = f\left( {{\alpha ^\mu },t_{k - 1}^\mu ,{\nabla _\mu }g\left( {\boldsymbol {p_{k+1}}} \right)} \right)$
		\item Calculate the importance measure ${\boldsymbol {\bar \omega}}_{k+1}= {\boldsymbol {\bar \omega}}_k - \left(\left(\nabla g\left( {{\boldsymbol p_{k+1}}} \right)\right)^T \odot \left( {{\boldsymbol p_{k+1}} - {\boldsymbol p_{k}}} \right)^T\right)^T$
	\end{enumerate}
	\textbf{end for}\\
	Normalize $\bar {\boldsymbol \omega}$ by (\ref{normalizedw}) and denote as ${ {\boldsymbol \omega} }$
\end{algorithm}

The proximal function $prox$ is defined and the proximal gradient is calculated by the soft threshold \cite{li2015accelerated}
	\begin{center}
$\begin{aligned}
  pro{x_{h,t}}\left( \boldsymbol p  \right) =& \arg \mathop {\min }\limits_{\boldsymbol z} \frac{1}{{2t}}\left\| {\boldsymbol z -\boldsymbol p} \right\|_2^2 + \lambda {\left\|\boldsymbol z  \right\|_1} \\
  = &{S_{\lambda t}}\left( \boldsymbol p  \right)
  \end{aligned}$
\end{center}
The soft threshold ${S_{\lambda t}}\left( \boldsymbol p \right)$ has an analytical solution \cite{dhingra2019the}:
\begin{equation}\label{softthresholdoperator}
  {\left[ {{S_{\lambda t}}} \right]_i} = \left\{ {\begin{array}{*{20}{l}}
{{p _i} - \lambda t,}&{{p _i} > \lambda t}\\
{0,}&{\left| {{p_i}} \right| \le \lambda t}\\
{{p_i} + \lambda t,}&{{p_i} < \lambda t}
\end{array}} \right.
\end{equation}
where $p_i$ is the $i$th element of $\boldsymbol p$.

According to (\ref{importance_mearure_v1}), the importance measure
is computed by
\begin{equation}\label{importance_mearurep_v2}
\bar{ \boldsymbol \omega } = \sum\limits_k \left(\left(\nabla g\left( {{\boldsymbol p_{k}}} \right)\right)^T \odot \left( {{\boldsymbol p_{k}} - {\boldsymbol p_{k-1}}} \right)^T \right)^T
\end{equation}
where $\odot$ denotes the Khatri-Rao product and $\boldsymbol p_{k}$ is the projection vector at $k$th iteration step. Accordingly,
each element of $\bar {\boldsymbol \omega}$ is normalized by
\begin{equation}\label{normalizedw}
{{ \omega }_i} = \max \left( {0,\frac{{{\bar \omega _i}}}{{{{\left( {\Delta {p_i}} \right)}^2} + \zeta }}} \right)
\end{equation}
where $\Delta {p_i}$ represents the total change, $1 \le i \le m$.  The solution is summarized in Algorithm \ref{APG}. 

The procedure of SPCA-SI is summarized in Algorithm \ref{SPCASItraining}.
For convenience, the optimal projection matrix and the importance measure are denoted as ${\boldsymbol P}_{\mathcal{M}_1}$ and ${\boldsymbol \Omega}_{\mathcal{M}_1}$, respectively.  

\begin{algorithm}[!bp]
	\caption{Procedure of SPCA-SI}\label{SPCASItraining}
	\begin{algorithmic}[1]
		\REQUIRE  data $\boldsymbol X$,  $l$ \\ 
		\ENSURE   The projection matrix $\boldsymbol P$, the importance measure $\boldsymbol \Omega$ \\
        \STATE    Initialize $\boldsymbol P^0 = \left[ {\begin{array}{*{20}{c}}
                 {{\boldsymbol p^0_1}}& \cdots &{{\boldsymbol p^0_l}}
          \end{array}} \right] = {I_{m,l}} $, $j=1$;
		\STATE    Scale $\boldsymbol X$ to zero mean and unit variance;
		\STATE    Let $\boldsymbol p_0 = {\boldsymbol p^0_j}$,  solve  (\ref{LAGRANGE})  by APG as summarized in Algorithm \ref{APG}. The gradients are calculated by (\ref{gradp1}-\ref{gradmu});
		\STATE   The optimal projection vector and importance measure are denoted as ${\boldsymbol p_j}$ and ${\boldsymbol \omega_j}$. Deflate $\boldsymbol X$ as $\boldsymbol X: = \boldsymbol X - \boldsymbol X \boldsymbol p_j {\boldsymbol p_j^T}$;
		\STATE  Let $j=j+1$, return to step 3  until $j>l$;  
        \STATE  ${\boldsymbol P = \left[ {\begin{array}{*{20}{c}}
                 {{\boldsymbol p_1}}& \cdots &{{\boldsymbol p_l}}
          \end{array}} \right]}$, ${\boldsymbol \Omega = \left[ {\begin{array}{*{20}{c}}
                 {{\boldsymbol \omega_1}}& \cdots &{{\boldsymbol \omega_l}}
          \end{array}} \right]}$.
	\end{algorithmic}
\end{algorithm}

\subsection{SPCA-SI for multiple modes}
When the  mode $\mathcal{M}_i$ ($i \ge 2$)  arrives, only   data $\boldsymbol X_i$ are available for training and data from previous modes are not retained.  SPCA-SI aims to learn the new mode while consolidating  the acquired knowledge of previous modes.
%
As shown in Fig. \ref{fig3}, we need to minimize the total loss of all modes, with the constraint that the loss functions of previous trained modes are unavailable. To alleviate catastrophic forgetting,  drastic changes to influential parameters in the past should be avoided. Therefore, a quadratic surrogate loss is introduced to approximate the total loss of previous modes \cite{zenke2017continual}.

The model of SPCA-SI is updated based on the current model and new data. For the  $j$th projection vector ($1\le j \le l$),
the objective is designed as:
\begin{equation}\label{originalaim2_v1}
  \begin{aligned}
\min \; &\left\| {\boldsymbol X_i - \boldsymbol X_i \boldsymbol p{\boldsymbol p^T}} \right\|_F^2 +  {\left( {\boldsymbol p - {\boldsymbol p_{\mathcal{M}_{i-1}}}} \right)^T} \bar {\boldsymbol \Omega} \left( \boldsymbol p - {\boldsymbol p_{\mathcal{M}_{i-1}}} \right)\\
 &
 +\lambda \left\|\boldsymbol p\right\|_1\\
s.t. \; &{\boldsymbol p^T}\boldsymbol p = 1
\end{aligned}
\end{equation}
where ${\boldsymbol p_{\mathcal{M}_{i-1}}}$ is the $j$th column of ${\boldsymbol P_{\mathcal{M}_{i-1}}}$, $\bar {\boldsymbol \Omega} =\gamma_i diag\left(  \boldsymbol \omega\right)$,
 $\boldsymbol \omega$ is the importance measure corresponding to ${\boldsymbol p_{\mathcal{M}_{i-1}}}$ and the $j$th column of ${{\mathord{\buildrel{\lower3pt\hbox{$\scriptscriptstyle\smile$}}
\over {\boldsymbol \Omega} } }_{\mathcal{M}_{i - 1}}}$,   ${{\mathord{\buildrel{\lower3pt\hbox{$\scriptscriptstyle\smile$}}
\over {\boldsymbol \Omega} } }_{\mathcal{M}_{i - 1}}} = \sum\nolimits_{r = 1}^{i - 1} {{{\boldsymbol \Omega} _{\mathcal{M}_r}}} $, and $\gamma_i$ is the weight which trades off previous versus current modes.
Similar to \cite{KirkpatrickOvercoming,zenke2017continual}, the additional regularization term is the quadratic surrogate loss, which makes the optimal parameters of current mode close  to the previous one, with a small loss.

Similarly, the augmented Lagrangian function is depicted as
\begin{equation}\label{originalaim2_v2}
  \begin{aligned}
J =& \left\| {\boldsymbol X_i - \boldsymbol X_i \boldsymbol p{\boldsymbol p^T}} \right\|_F^2+\lambda \left\|\boldsymbol p\right\|_1 + \mu \left({\boldsymbol p^T}\boldsymbol p - 1\right)^2  \\
&+{\left( {\boldsymbol p - {\boldsymbol p_{\mathcal{M}_{i-1}}}} \right)^T} \bar {\boldsymbol \Omega} \left( \boldsymbol p - {\boldsymbol p_{\mathcal{M}_{i-1}}} \right)\\
\end{aligned}
\end{equation}
The smooth part $g\left( \boldsymbol p \right)$ and the corresponding gradient are
\begin{equation}\label{smoothg2}
\begin{aligned}
  g\left( \boldsymbol p \right)=&\left\| {\boldsymbol X_i - \boldsymbol X_i \boldsymbol p \boldsymbol p^T} \right\|_F^2 +  {\left( {\boldsymbol p - {\boldsymbol p_{\mathcal{M}_{i-1}}}} \right)^T} \bar {\boldsymbol \Omega} \left( \boldsymbol p - {\boldsymbol p_{\mathcal{M}_{i-1}}} \right)\\
&+\mu {\left( {\boldsymbol p^T}\boldsymbol p - 1\right)^2}
  \end{aligned}
\end{equation}
\begin{equation} \label{gradp2}
\begin{aligned}
 {\nabla _{\boldsymbol p}}g\left( \boldsymbol p \right) =  &\boldsymbol p{\boldsymbol p^T} \boldsymbol G_i \boldsymbol p + \boldsymbol G_i \boldsymbol p{\boldsymbol p^T}\boldsymbol p
 - 2\boldsymbol G_i \boldsymbol p+2 \bar {\boldsymbol \Omega} \left( \boldsymbol p - {\boldsymbol p_{\mathcal{M}_{i-1}}} \right)
 \end{aligned}
\end{equation}
where $\boldsymbol G_i=2\left( {{\boldsymbol X_i^T}\boldsymbol X_i + \mu \boldsymbol  I}\right)$.

The solution can refer to Algorithm \ref{SPCASItraining}. The optimization problem is (\ref{originalaim2_v2}), and the gradients are calculated by  (\ref{gradp2}) and (\ref{gradmu}).
The optimal projection matrix  and  importance measure are denoted as $\boldsymbol P_{\mathcal{M}_i}$ and   $\boldsymbol \Omega_{\mathcal{M}_i}$, respectively.


\section{Monitoring model and Discussion}\label{sec:statistic}

\subsection{Monitoring statistics}

Two statistics are designed to monitor the operating condition. 
To enhance the monitoring performance for previous modes, the partial covariance information of last mode is adopted  to calculate  $T^2$ statistic.
For  mode $\mathcal{M}_i $  ($i \ge 1$),
\begin{equation}\label{T2}
{T^2} = {\boldsymbol x^T}\boldsymbol P_{\mathcal{M}_{i}}{\boldsymbol \Xi_{\mathcal{M}_{i} } ^{ - 1}}{\boldsymbol P_{\mathcal{M}_{i}}^T} \boldsymbol x
\end{equation}
where
$\boldsymbol x \in \boldsymbol X_i$, $\boldsymbol P_{\mathcal{M}_{i}}$ is the projection matrix,
$\boldsymbol \Xi_{\mathcal{M}_{i}}  = \boldsymbol P_{\mathcal{M}_{i}}^T \left( \eta  \frac{{{\boldsymbol X_i^T}\boldsymbol X_i}}{{N_i - 1}} + (1 - \eta )\boldsymbol P_{\mathcal{M}_{i-1}} {\boldsymbol \Xi}_{\mathcal{M}_{i-1} } {\boldsymbol P_{\mathcal{M}_{i-1}}^T}\right)\boldsymbol P_{\mathcal{M}_{i}}$, and $N_i$ is the number of samples, $\eta$ trades off the previous versus current modes with $0 \le \eta \le 1$.
$\boldsymbol P_{\mathcal{M}_{i-1}}$ and ${\boldsymbol \Xi}_{\mathcal{M}_{i-1}}$ are acquired from mode $\mathcal{M}_{i-1}$, 
which represent the information of previous modes without storing the original data.
  When $i=1$, let $\eta =1$, $\boldsymbol \Xi_{\mathcal{M}_{1}}= \frac{{{\boldsymbol X_1^T}\boldsymbol X_1}}{{N_1 - 1}} $. If $i>1$, $\eta $ is estimated by the importance of previous modes.
Correspondingly, the squared prediction error (SPE) is calculated by
\begin{equation}\label{SPE}
SPE = {\boldsymbol x^T}\left( {I - \boldsymbol P_{\mathcal{M}_{i}}{\boldsymbol P_{\mathcal{M}_{i}}^T}} \right) \boldsymbol x
\end{equation}


The thresholds are calculated by kernel density estimation (KDE) \cite{zhang2019an} and the confidence level is $99\%$. Once one statistic is beyond its threshold, a fault is detected. The training procedure is summarized in Algorithm \ref{SPCASItrainingprocedure}.

\begin{algorithm}[!tbp]
	\caption{Off-line training phase of SPCA-SI}\label{SPCASItrainingprocedure}
	\begin{algorithmic}[1]
   \STATE   For the mode $\mathcal{M}_1$, collect data  $\boldsymbol X_1$;
        \STATE    Normalize $\boldsymbol X_1$ to zero mean and unit variance;
        \STATE    Perform  traditional PCA on $\boldsymbol X_1$ and calculate the number of principal components $l$;
        \STATE   Solve the  issue (\ref{LAGRANGE}) by  Algorithm \ref{SPCASItraining}, acquire $\boldsymbol P_{\mathcal{M}_1}$ and  ${\boldsymbol \Omega}_{\mathcal{M}_1}$. The gradients  are calculated by (\ref{gradp1}-\ref{gradmu}); 
        \STATE Calculate statistics by (\ref{T2}-\ref{SPE}) and 
 thresholds  by KDE;
		\STATE   For the mode $\mathcal{M}_i$ ($i \ge 2$), collect data  $\boldsymbol X_i$;
        \STATE   Scale $\boldsymbol X_i$ to zero mean and unit variance;
      \STATE  Solve the  issue (\ref{originalaim2_v2}) by  Algorithm \ref{SPCASItraining}, acquire $\boldsymbol P_{\mathcal{M}_i}$ and  ${\boldsymbol \Omega}_{\mathcal{M}_i}$. The gradients  are calculated by  (\ref{gradmu}) and  (\ref{gradp2}) ;
        \STATE Calculate statistics by (\ref{T2}-\ref{SPE}) and 
thresholds by KDE.

	\end{algorithmic}
\end{algorithm}

\subsection{Discussion}

\subsubsection{Parameter setting}

SPCA-SI  has three regularization parameters. $\lambda$ affects the sparsity and SPCA is transformed to PCA when $\lambda=0$. Thus, SPCA-SI is equivalent to PCA-SI. Similarly, PCA-SI offers continual learning ability but the parameters are not sparse.

Then, we explain how the continual learning ability is influenced by $\gamma_i$ and $\eta$.  Two extreme cases are given as an example.  When $\gamma_i =0$ and $\eta=1$, SPCA-SI is equivalent to traditional SPCA and information of previous modes is forgotten catastrophically (similar to Fig. \ref{fig2}). When $\gamma_i  \to \infty $ and $\eta=0$, the information of  mode $\mathcal{M}_1$ is completely preserved while the knowledge of  subsequent modes is not learned. In other cases, information of different modes is memorized and beneficial to monitor multiple modes simultaneously.


\subsubsection{Computational complexity}

The computational complexity focuses on Algorithms \ref{APG} and \ref{SPCASItraining}.  $k_{total}$ is the total number of iterations. Here we use $flam$ to reflect the complexity.
The calculation of $\boldsymbol G_i$ needs $\frac{1}{2}N_im^2+2m$ flam.  For mode $\mathcal{M}_1$, 
 $10m^2+38m+14$ flam is required for each iteration in Algorithm \ref{APG}.
Thus, the computational complexity is $(10m^2+38m+14)k_{total}+ l(\frac{1}{2}N_1m^2+3N_1m+2m)$ flam for training.
For mode $\mathcal{M}_i$ ($i\ge 2$),  
Algorithm \ref{APG} needs $10m^2+50m+14$ flam in total for each iteration.  The training  phase needs $(10m^2+50m+14)k_{total}+l(\frac{1}{2}N_im^2+3N_im+2m)$ flam. 


%
%
%
%

%

%
%
%

\section{Case study}\label{sec:4}
This section adopts two case studies to illustrate the effectiveness of SPCA-SI.  Recursive PCA (RPCA) \cite{li2000recursive} and improved mixture of probabilistic PCA (IMPPCA) \cite{zhang2019an} are employed for comparison. For IMPPCA,
 the mode is automatically identified by membership degree. The fault detection rate (FDR) and false alarm rate (FAR) are  employed to evaluate the monitoring performance.

\begin{table}[!bp]
	\begin{center}
		\caption{Comparative scheme for numerical case}\label{comparative-numerical}
		\footnotesize
		\begin{tabular}{c c c c c}
			\hline
			& Methods &  \makecell{Training \\sources}  & \makecell{Model\\ label} & \makecell{Testing \\sources}  \\
			\hline
			Situation 1 & SPCA       & Mode 1                    &  A      & Mode 1         \\
			Situation 2 & SPCA-SI  & Model A+Mode 2    &  B      & Mode 2       \\
			Situation 3 & SPCA-SI  &  -                              &   B      & Mode 1     \\
			Situation 4 & SPCA       &  Mode 2                  &  C       & Mode 2       \\
			Situation 5 & SPCA       &  -                             &  C       & Mode 1      \\
			Situation 6 & RPCA      & Modes 1,2              &  D      & Mode 1      \\
			Situation 7 & RPCA      & -                              &   D      & Mode 2    \\
			Situation 8 & IMPPCA  & Modes 1,2              &   E     & Mode 1     \\
			Situation 9 & IMPPCA  &  -                              &   E      & Mode 2     \\
			\hline
		\end{tabular}
	\end{center}
\end{table}

\subsection{Numerical case}
The following numerical case is adopted:
\begin{center}
$ \left[ {\begin{array}{*{20}{c}}
	{{x_1}}\\
	{{x_2}}\\
	{{x_3}}\\
	{{x_4}}\\
	{{x_5}}\\
	{{x_6}}\\
	{{x_7}}\\
	{{x_8}}
	\end{array}} \right]  = \left[ {\begin{array}{*{20}{c}}
	{0.55}&{0.82}&{0.94}\\
	{0.23}&{0.45}&{0.62}\\
	{ - 0.61}&{0.62}&{0.41}\\
	{0.49}&{0.79}&{0.89}\\
	{0.89}&{ - 0.92}&{0.06}\\
	{0.76}&{0.74}&{0.35}\\
	{0.46}&{0.28}&{0.81}\\
	{ - 0.02}&{0.41}&{0.01}
	\end{array}} \right]\left[ {\begin{array}{*{20}{c}}
	{{s_1}}\\
	{{s_2}}\\
	{{s_3}}
	\end{array}} \right] + \boldsymbol e $
 \end{center}
  where the  noise $\boldsymbol e$ follows Gaussian distribution  with $ e_i \sim \mathbb{N}(0,0.001), i=1,\cdots,8$.   Sequential data are generated successively from two modes: \\
 $\bullet$   Mode 1: $s_1 \sim \mathbb{U}([-10,-9.7])$, $ s_2 \sim \mathbb{N}(-5,1)$, and $s_3 \sim \mathbb{U}([2,3])$; 	\\
 $\bullet$   Mode 2: $s_1 \sim \mathbb{U}([-6,-5.7])$, $ s_2 \sim \mathbb{N}(-1,1)$, and $s_3 \sim \mathbb{U}([3,4.2])$;  	\\
 \noindent where $\mathbb{U}([-10,-9.7])$ represents the uniform distribution between $-10$ and $-9.7$, and so on.

We generate 1000 normal samples to train the monitoring model and  1000 samples for fault detection, including  the first 500 normal samples  and the last 500 faulty samples  through the following  scenarios:\\
 $\bullet$  Fault 1: step fault of  $x_3$, $x_3 = x_3^*+0.08$; \\ 
 $\bullet$  Fault 2: step fault of  $x_6$, $x_6 = x_6^*+0.08$; \\ 
 $\bullet$  Fault 3: slope drift of $x_1$, $x_1 = x_1^*+0.001(k-500)$;    

\noindent where $500 \le k \le 1000$, $x_1^*$, $x_3^*$ and $x_6^*$ are normal.

The simulation scheme is designed to illustrate the effectiveness and superiorities of  SPCA-SI, as summarized in Table \ref{comparative-numerical}. Note that `-' indicates that there is no need to retrain the model and the current monitoring model is  adopted for fault detection.
The first five situations are utilized to verify that  SPCA-SI alleviates the catastrophic forgetting of traditional SPCA and furnishes self-learning ability. Specifically, the results of Situations 1 and 4 testify the effectiveness of SPCA for a mode. The monitoring results of Situations 2 and 3 are used to  prove that SPCA-SI is able to monitor two modes simultaneously by a continually updated model, which assimilates new data when a new mode appears. Situation 5 is designed to show that SPCA fails to monitor the previous mode and the learned knowledge is overwritten by new information.
For situations 6-9, RPCA and IMPPCA are compared to illustrate the superiorities of  SPCA-SI  further.

\begin{figure*}[!htbp]
	\centering
\subfigure{\label{fault1-2}}\addtocounter{subfigure}{-2}
\subfigure
	{\subfigure[Situation 2]{\includegraphics[width=0.235\textwidth]{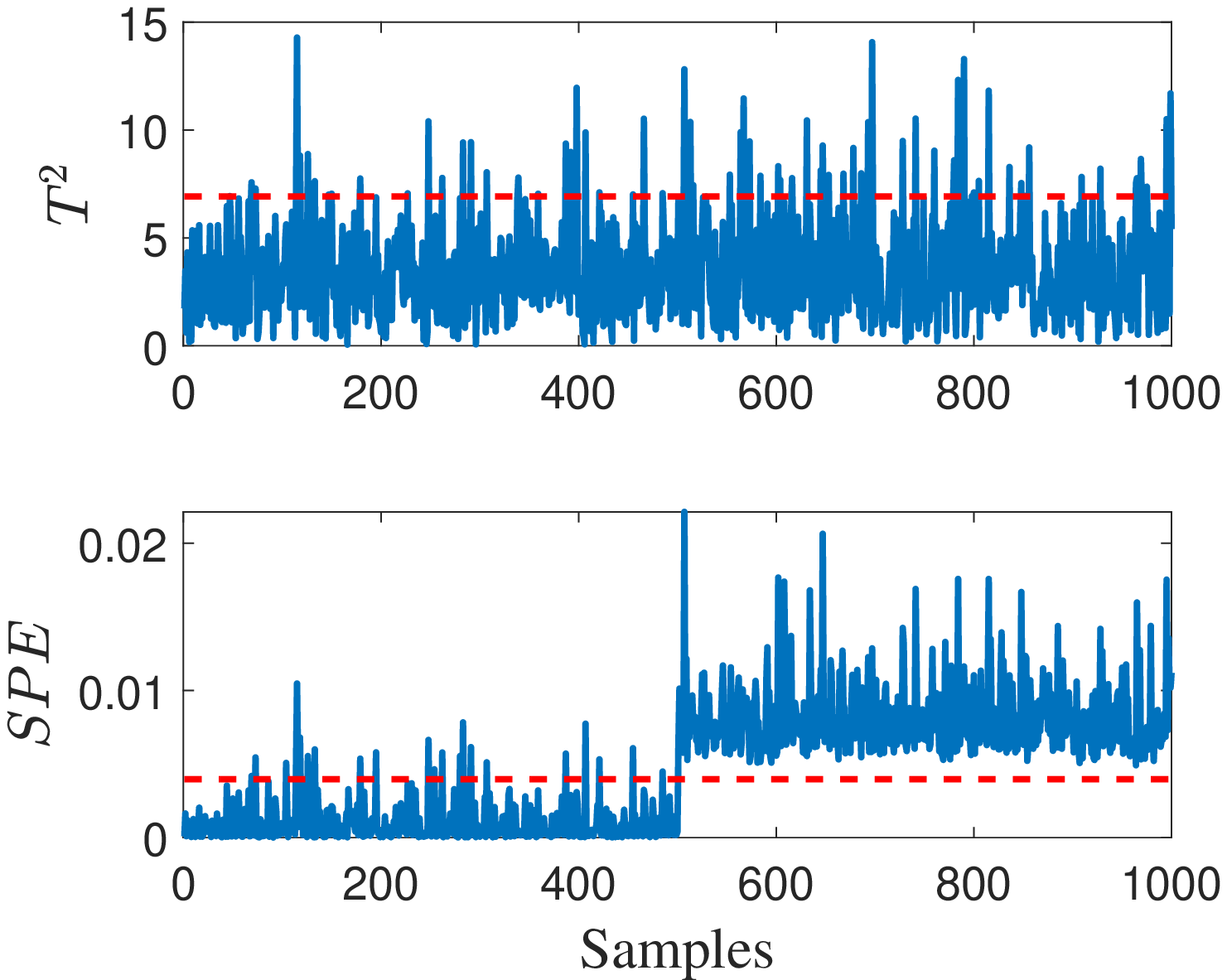} }}
       \subfigure{\label{fault1-3}}\addtocounter{subfigure}{-2}
	\subfigure
	{\subfigure[Situation 3]{\includegraphics[width=0.235\textwidth]{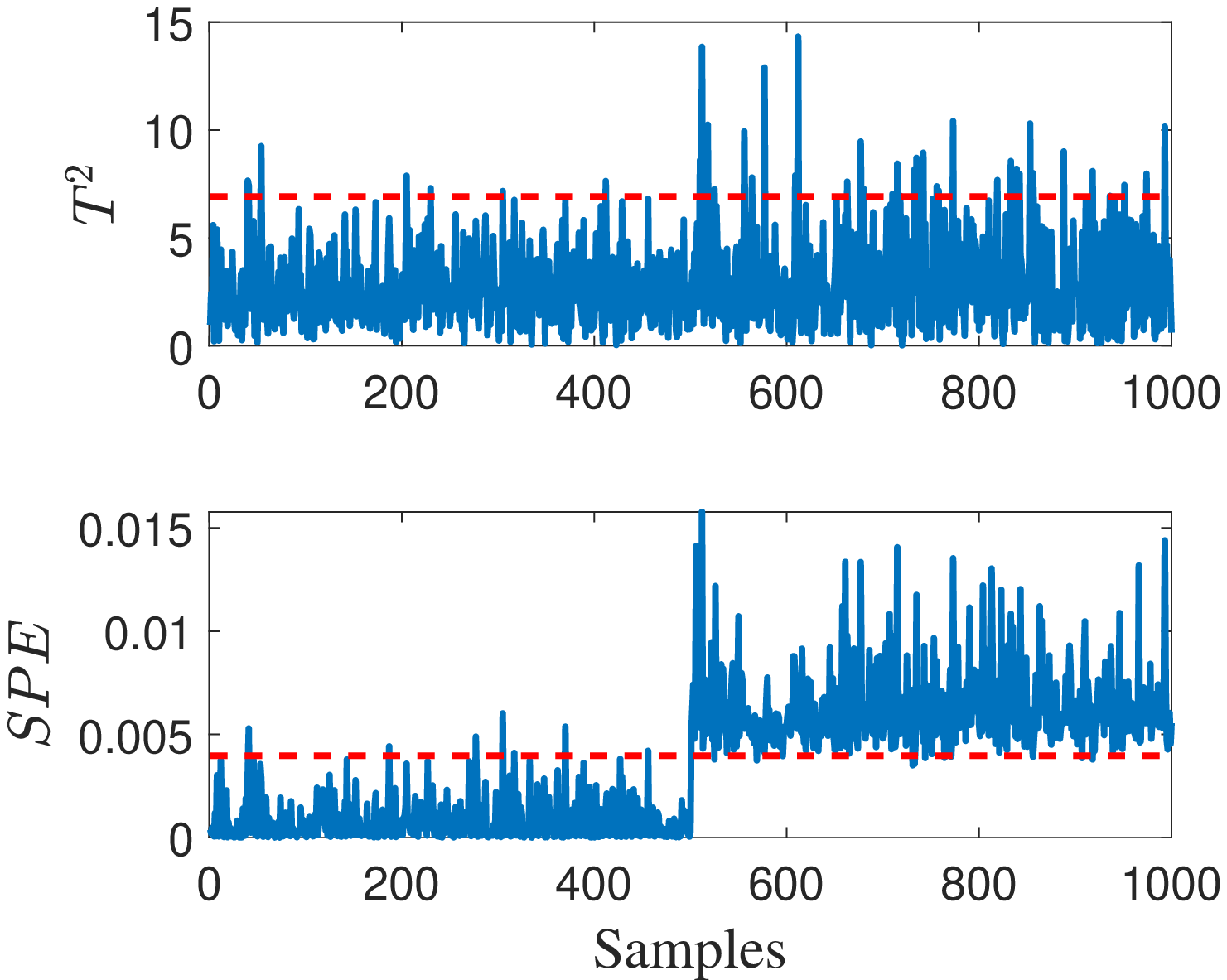}}}
	%
	\subfigure{\label{fault1-4}}\addtocounter{subfigure}{-2}	
	\subfigure
	{\subfigure[Situation 4]{\includegraphics[width=0.235\textwidth]{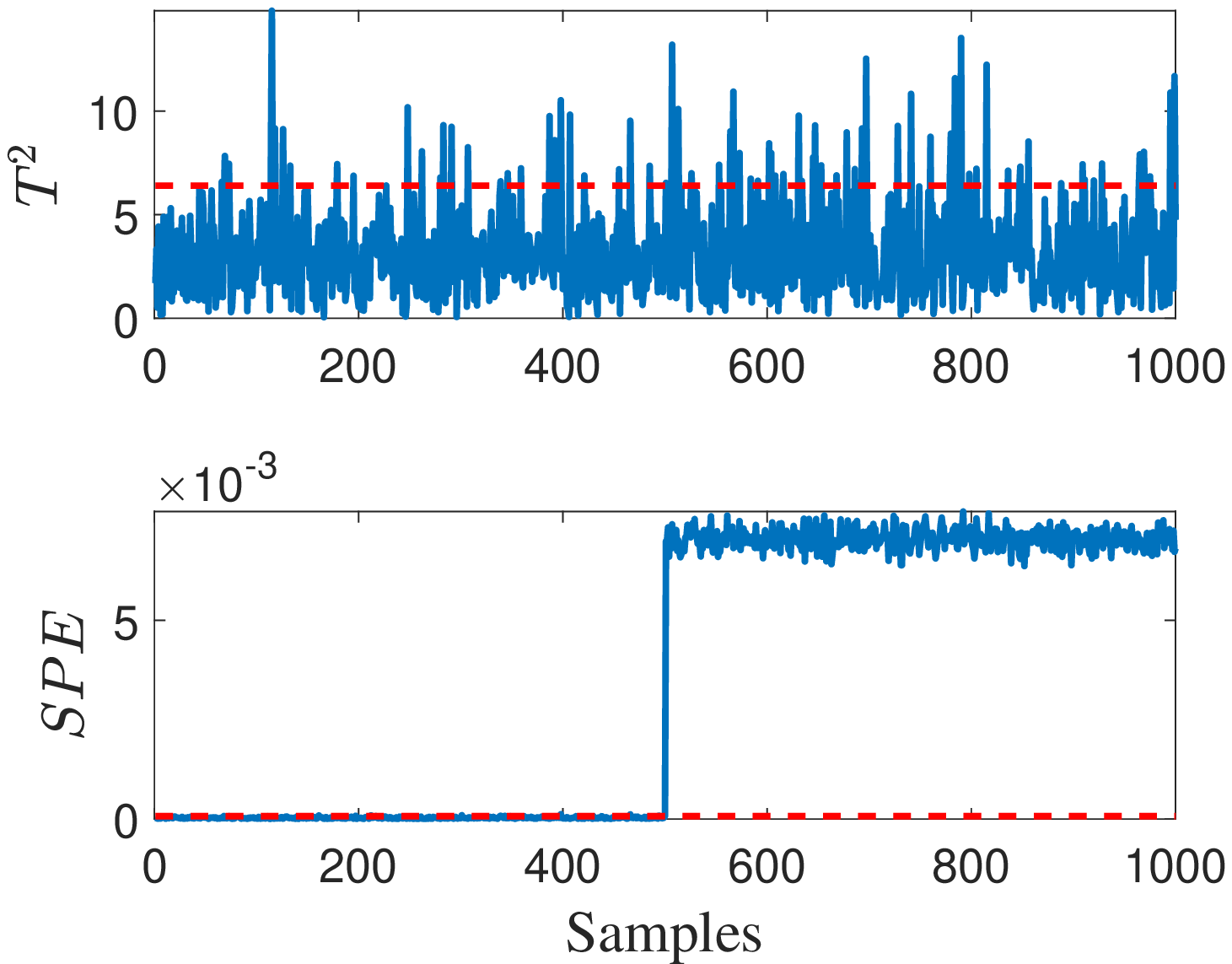}}}
\subfigure{\label{fault1-5}}\addtocounter{subfigure}{-2}
    	\subfigure
    	{\subfigure[Situation 5]{\includegraphics[width=0.235\textwidth]{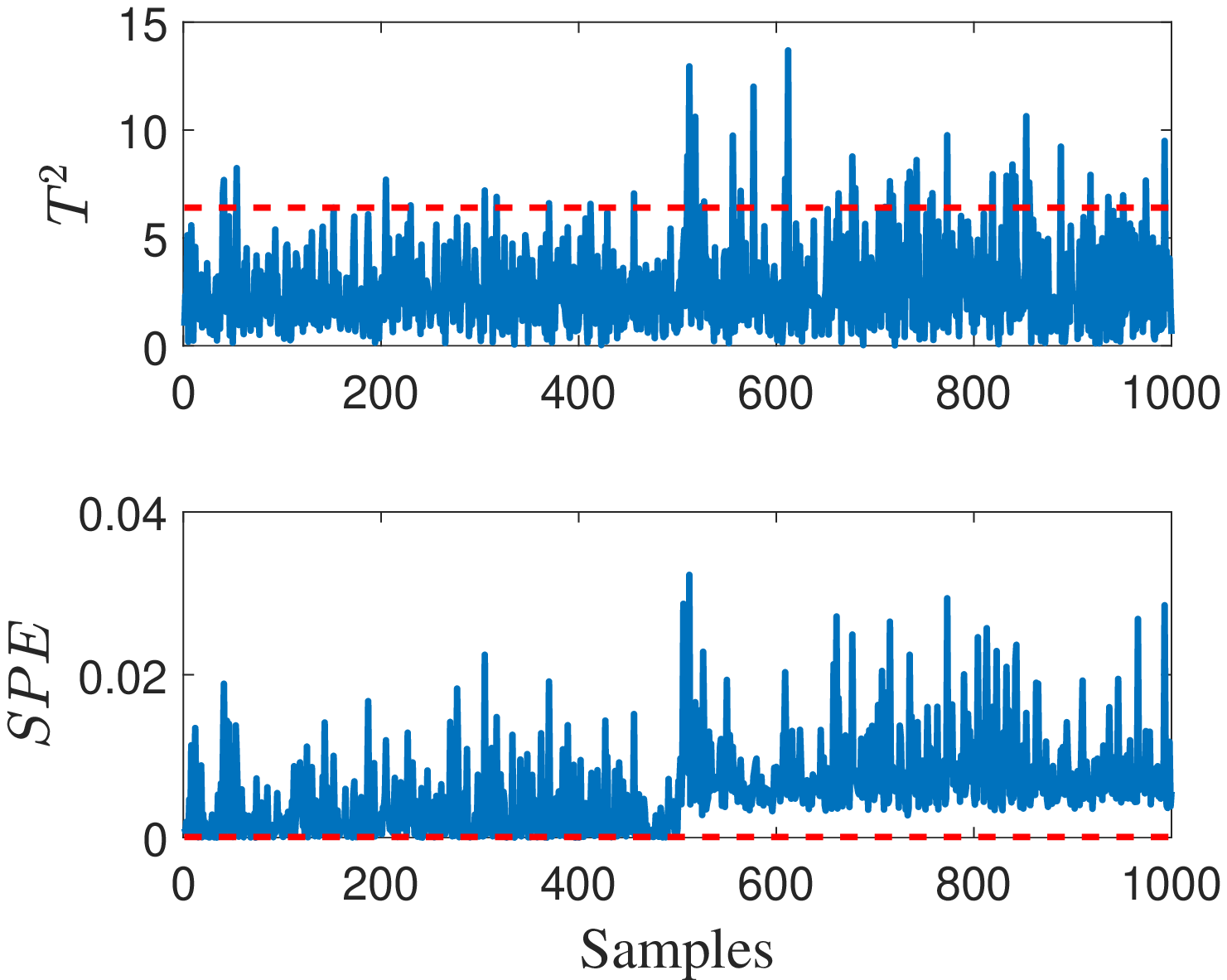}}}
    	\subfigure{\label{fault1-6}}\addtocounter{subfigure}{-2}
    	\subfigure
    	{\subfigure[Situation 6]{\includegraphics[width=0.235\textwidth]{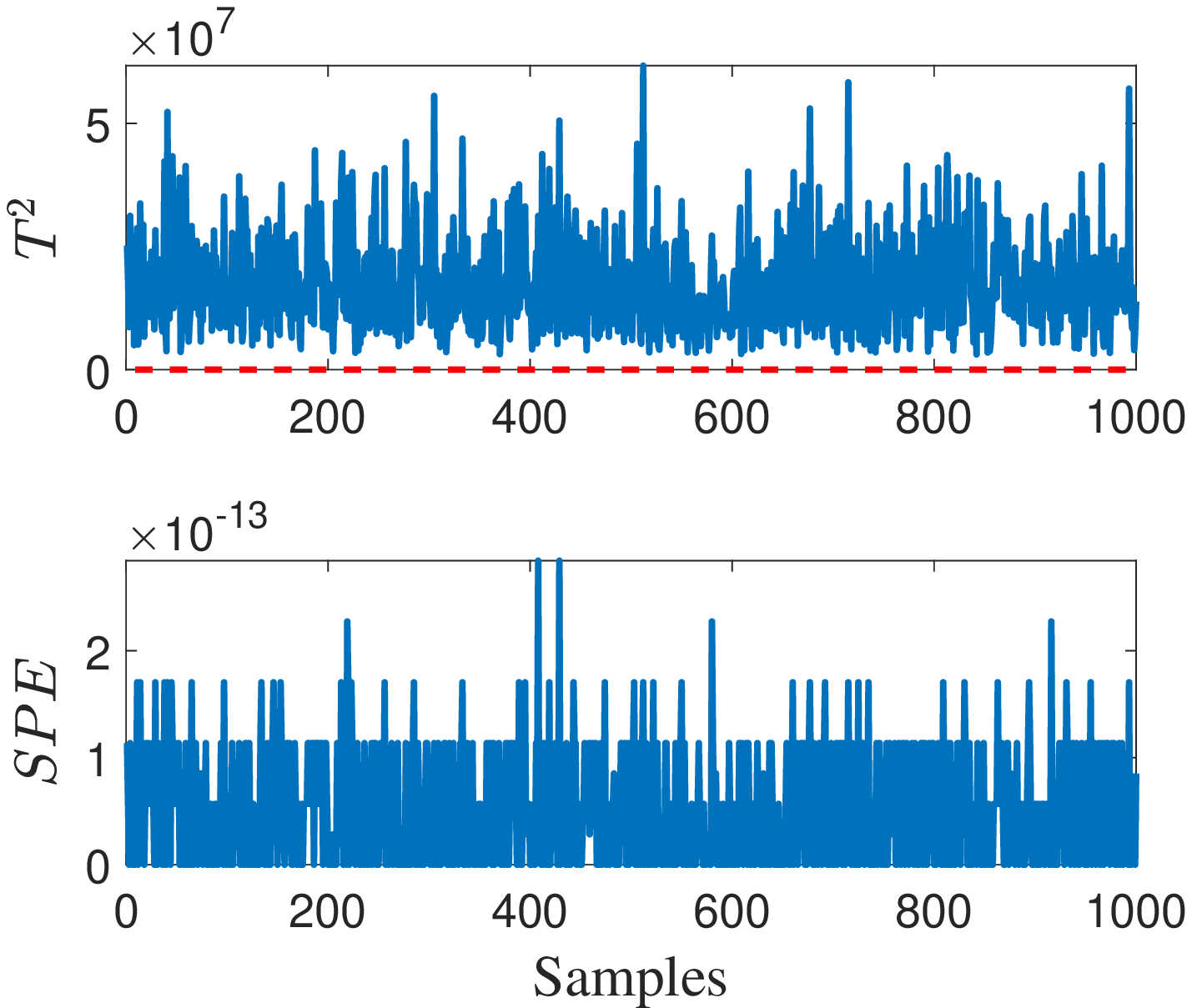}}}
    	\subfigure{\label{fault1-7}}\addtocounter{subfigure}{-2}
    	\subfigure
    	{\subfigure[Situation 7]{\includegraphics[width=0.225\textwidth]{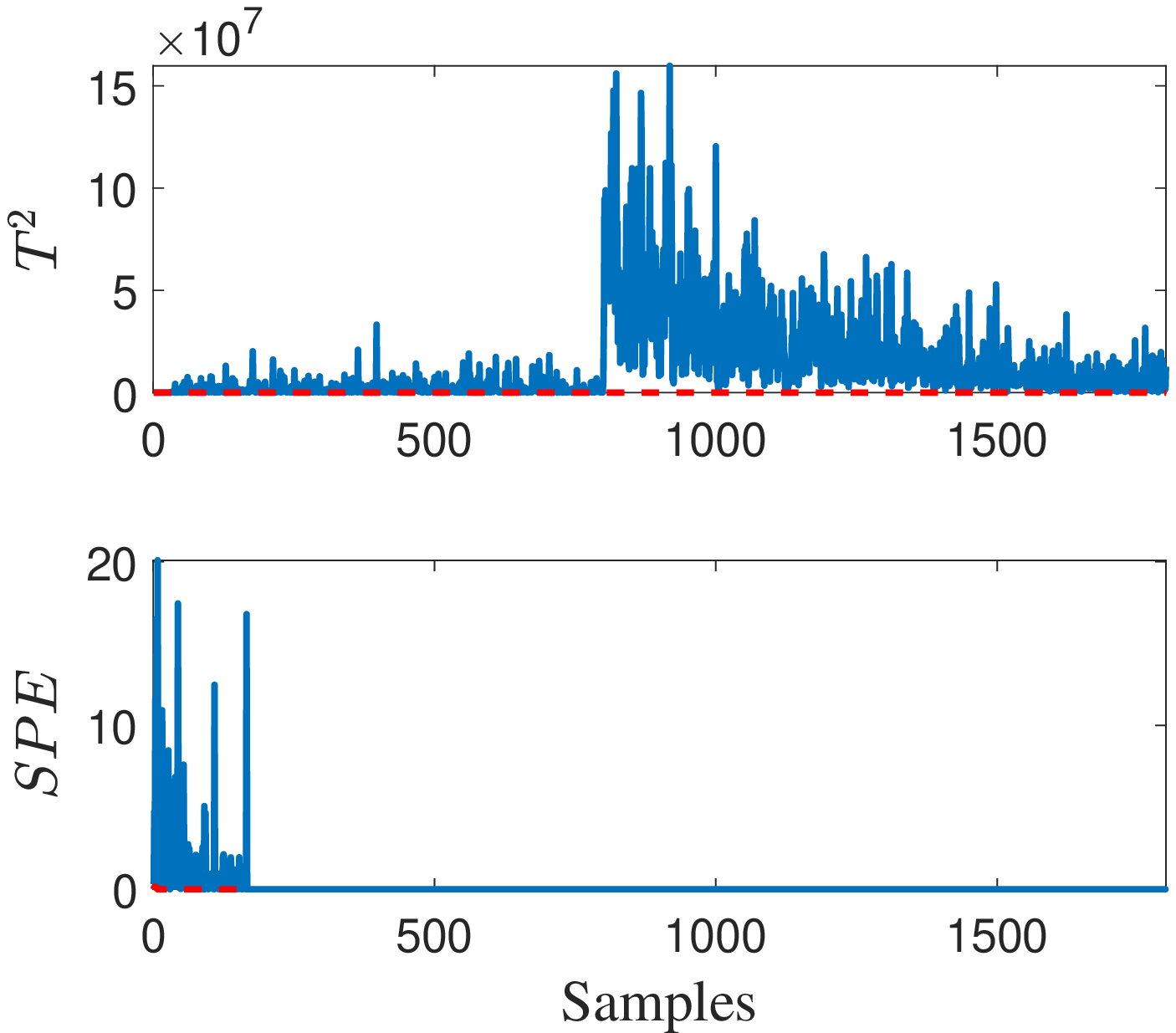}}}
    \subfigure{\label{fault1-8}}\addtocounter{subfigure}{-2}
    	\subfigure
    	{\subfigure[Situation 8]{\includegraphics[width=0.238\textwidth]{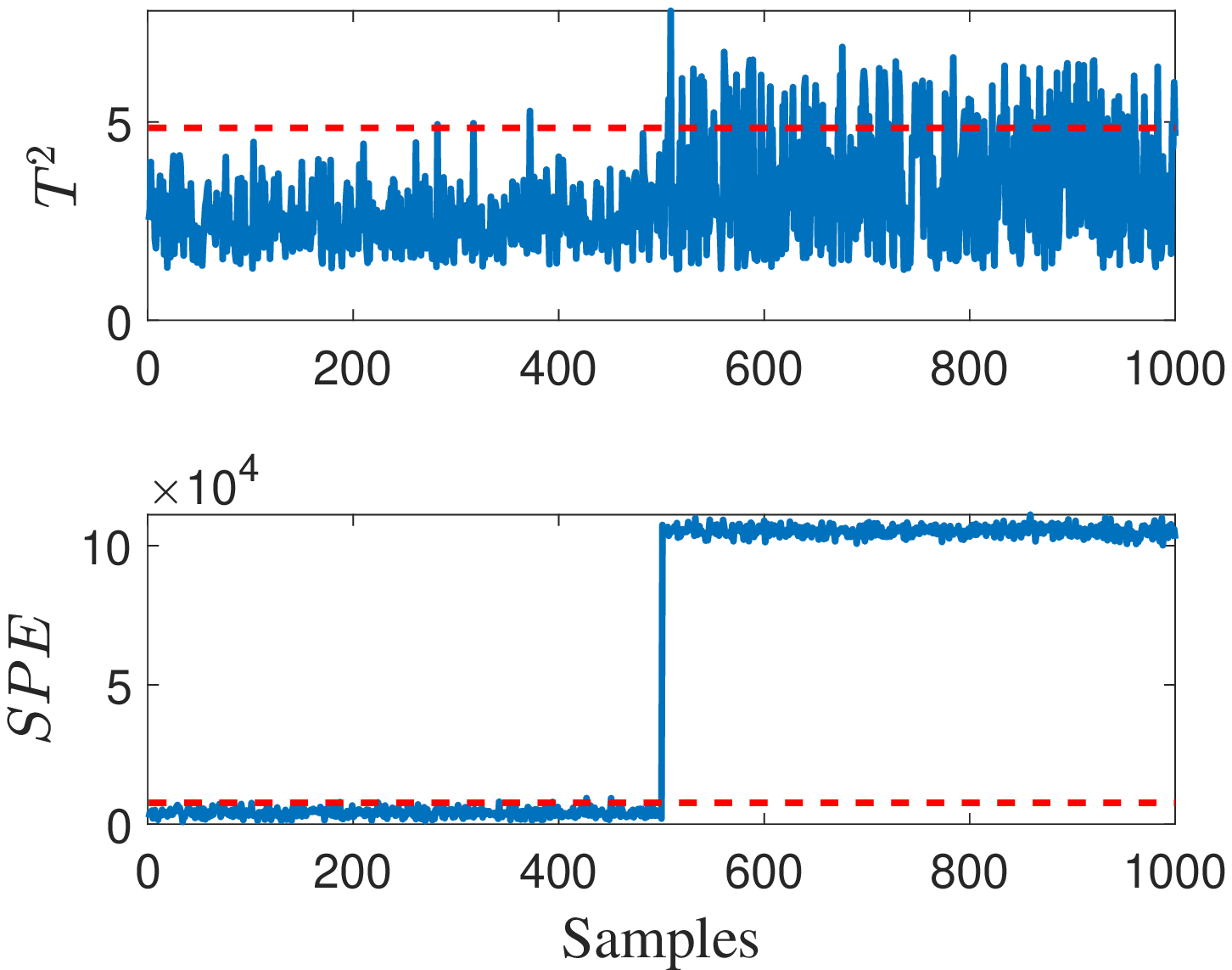}}}
        \subfigure{\label{fault1-9}}\addtocounter{subfigure}{-2}
    \subfigure
    	{\subfigure[Situation 9]{\includegraphics[width=0.238\textwidth]{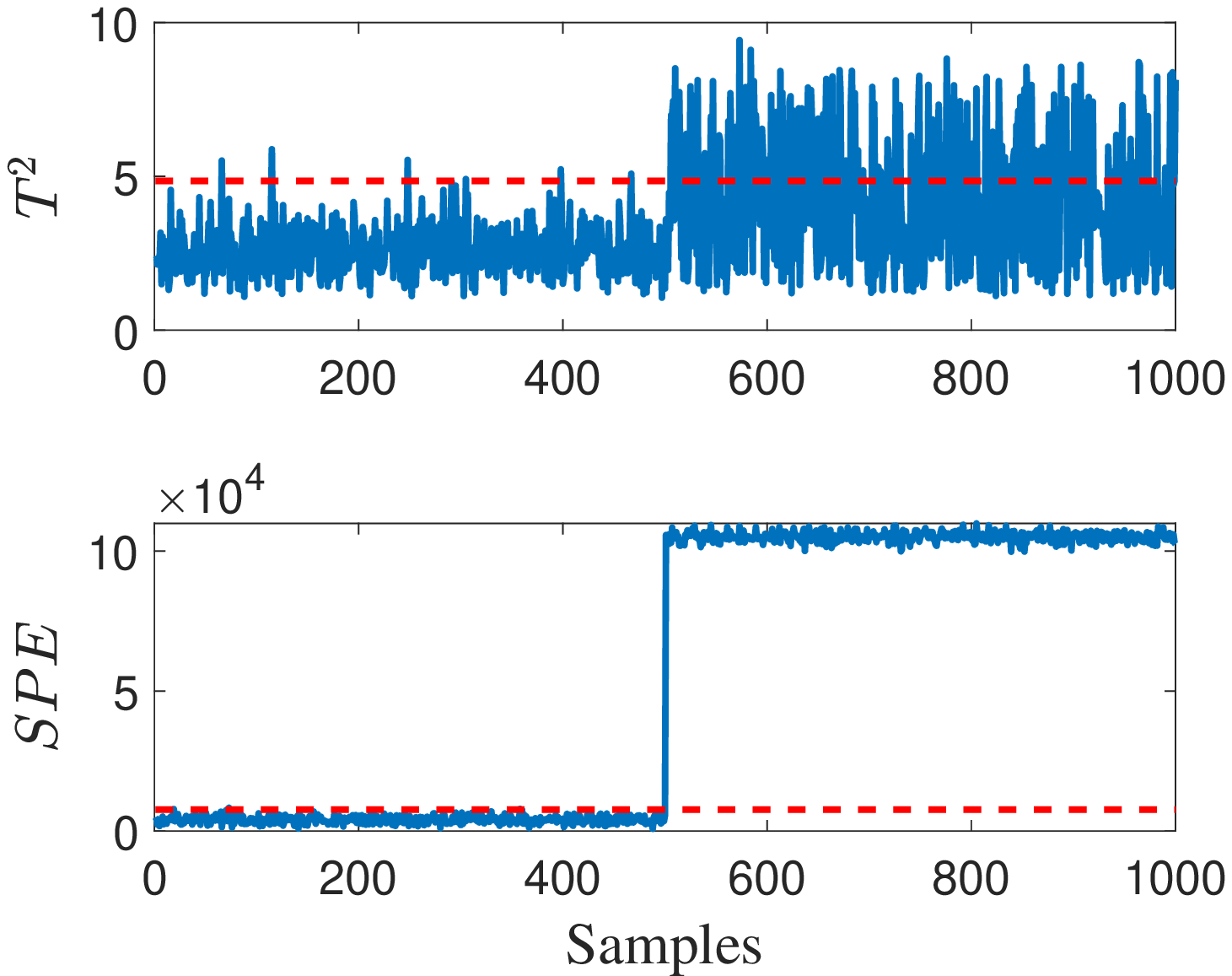}}}
	\centering
	\caption{Monitoring charts of Fault 1} \label{case1}
\end{figure*}

\begin{table}[!tp]
  	\begin{center}
  		\caption{FDR ($\%$) and FAR ($\%$) for numerical case}\label{Table1-comparativeresult}
  		\footnotesize
  		\begin{tabular}{c c c c c c c}
  			\hline
       Fault type &  \multicolumn{2}{c}{Fault 1} &  \multicolumn{2}{c}{Fault 2}  &  \multicolumn{2}{c}{Fault 3}\\
        \cmidrule(r){1-1}  \cmidrule(r){2-3}  \cmidrule(r){4-5} \cmidrule(r){6-7}
        Indexes     &   {FDR}    & {FAR} &   {FDR}    & {FAR} &   {FDR}    & {FAR}\\
        \hline
        Situation 1 & 100    & 7.4   & 100    & 2.6   & 96.8  & 0\\
        Situation 2 & 100    & 6.6   & 100    & 2.2   & 91.0  & 0\\
        Situation 3 & 98.6   & 2.4   & 99.6   & 1.0   & 90.6  & 4.6 \\
        Situation 4 & 100    & 8.4   & 100    & 2.6   & 96.4  & 0 \\
        Situation 5 & 100    & 93.4  & 100    & 90    & 98.8  & 65.2 \\
        Situation 6 & 100    & 98.4  & 100    & 98.4  & 100   & 98.4\\
        Situation 7 & 100    & 33.2  & 100    & 100   & 100   &100 \\
        Situation 8 & 100    & 2.2   & 100    & 2.6   & 95.2  & 1.4 \\
        Situation 9 & 100    & 2.2   & 100    & 1.8   & 96.0  &2.0\\
    	\hline
  		\end{tabular}
  	\end{center}
  \end{table}

Take Fault 1 as an example to interpret the monitoring consequences in detail, as described in Fig. \ref{case1}. As the monitoring charts for Mode 1 and Mode 2 are similar for SPCA, the simulation chart of Situation 1 is not listed. In Figs. \ref{fault1-2}-\ref{fault1-3}, SPCA-SI enables to detect the fault in Modes 1 and 2 accurately by model B, which is updated based on the existing model A and data from Mode 2. The model C fails to detect the fault in Mode 1, as depicted in Fig. \ref{fault1-5}. The FDR is $100\%$ and FAR is $93.4\%$, which indicates that the learned knowledge of Mode 1 is forgotten when training the model C (similar to Fig. \ref{fig2}).
According to Situations 1-5,  SPCA-SI alleviates the catastrophic forgetting of SPCA and provides self-learning ability for successive modes. For Situation 6-7, RPCA is unable to track the system changes and distinguish the novelty from normality. In Figs. \ref{fault1-8}-\ref{fault1-9}, IMPPCA  detects the fault accurately and the FDRs are $100\%$.

The monitoring results of three faults are summarized in Table \ref{Table1-comparativeresult}.  The analysis aforementioned is equally applied to Fault 2 and Fault 3.  Traditional SPCA forgets the significant features of previous modes when training the same model sequentially and fails to deliver prominent performance. SPCA-SI is capable of monitoring two modes simultaneously based on a continually updated model and settles the catastrophic forgetting of SPCA. RPCA fails to distinguish between the normal modes and faults.  IMPPCA detects the fault accurately and  needs to be retrained from scratch when new modes appear. Thus, it consumes much more storage space and computational resource than SPCA-SI. Overall, SPCA-SI with self-learning ability is superior to IMPPCA and RPCA, as it can monitor multiple modes accurately, and the computation and storage resources are saved in the long run.

 \begin{figure}[!bp]
 	\centering
 	\includegraphics[width=0.28\textwidth]{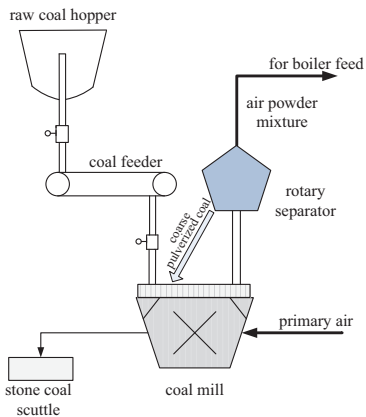}
 	\caption{Schematic diagram of the coal pulverizing system}
 	\label{fig_benchmark}
 \end{figure}

\subsection{Pulverizing system process monitoring}
 The 1000-MW ultra-supercritical thermal power plant is increasingly popular due to potential economic benefits and low pollution \cite{zhang2020multimode}. This paper investigates the coal pulverizing system in Zhoushan Power Plant in China, which provides high quality pulverized coal for boiler. As depicted in Fig. \ref{fig_benchmark}, it contains coal feeder, coal mill, rotary separator, raw coal hopper and stone coal scuttle. In practical systems, the types of coal and unit load would change frequently, thus generating successive operating modes.

%

In this paper, we focus on two typical faults, namely, abnormality from outlet temperature (Fault 4) and rotary separator (Fault 5). The  information is summarized in Table \ref{Table_informationZS}.  Note that the number of training samples and testing samples are shorted for NoTrS and NoTeS, respectively.  Nine key variables are selected by prior knowledge.
Assume that when new modes appear, the system operates under the normal condition at the preliminary stage.

To illustrate the effectiveness and self-learning ability of SPCA-SI, 17 situations are designed in Table \ref{Table1-comparative2}. Three different modes are considered in this case. Similar to the numerical case, Situations 1, 4 and 9 are utilized to illustrate the effectiveness of SPCA for a mode. Situations 2, 3, 6, 7 and 8 are designed to verify that SPCA-SI can monitor several modes simultaneously, where new data are assimilated while preserving the learned knowledge.
Situations 5, 10 and 11 are employed to show the catastrophic forgetting issue of SPCA. For Situations 12-14, RPCA is adopted to track the successive modes. Take  Situation 12 as an example, Modes 1-3 appear sequentially and Mode 1 occurs again, and the fault occurs in Mode 1.
For Situations 15-17, IMPPCA is adopted to monitor three modes and the training data are required  to be complete. 

\begin{table}[!tp]
	\begin{center}
		\caption{Data information of the coal pulverizing system}\label{Table_informationZS}
		\renewcommand\arraystretch{1.2}
		\small
		\begin{tabular}{c  c c c c} 
			\hline
			\makecell{Fault\\ type}  &  \makecell{Mode\\ number} & NoTrS  &NoTeS &\makecell{Fault\\ location}  \\
			\hline
			\multirow{3}{*}{Fault 4}
			& \makecell{$1$}   &2160  &2880 & 909  \\ 
			& \makecell{ $2$} & 1080  &1080 & 533   \\ 
			& \makecell{ $3$}  &1440  &1440 & 626  \\
			\hline		
			\multirow{3}{*}{Fault 5 }
			&\makecell{ $1$}  & 2880  &1080 & 806  \\ 
			& \makecell{ $2$}  & 720  &720 & 352  \\ 
			& \makecell{ $3$}  & 2880 &2160 & 134 \\
			\hline
		\end{tabular}
	\end{center}
\end{table}

\begin{table}[!tbp]
  	\begin{center}
  		\caption{Simulation scheme for the pulverizing system}\label{Table1-comparative2}
  		\footnotesize
  		\begin{tabular}{c c c c c}
  			\hline
  			            & Methods &  \makecell{Training\\sources}   & \makecell{ Model\\ label} & \makecell{Testing \\sources}   \\
  			\hline
            Situation 1 & SPCA          & Mode 1                    &  A    & Mode 1    \\
            Situation 2 & SPCA-SI     & Model A+Mode 2    &  B    & Mode 2      \\
            Situation 3 & SPCA-SI     & -                              & B     & Mode 1    \\
            Situation 4 & SPCA          &  Mode 2                   &  C   & Mode 2    \\
            Situation 5 & SPCA          &       -                        &  C    & Mode 1   \\
            Situation 6 & SPCA-SI     & Model B+Mode 3     &  D   & Mode 3    \\
            Situation 7 & SPCA-SI     &     -                           & D   & Mode 1   \\
            Situation 8 & SPCA-SI     &     -                           & D     & Mode 2   \\
            Situation 9 & SPCA          & Mode 3                    &  E    & Mode 3   \\
            Situation 10& SPCA         &     -                           &  E   & Mode 1  \\
            Situation 11& SPCA         &     -                           &  E   & Mode 2  \\
            Situation 12& RPCA        & Modes 1, 2, 3           & F     & Mode 1   \\
            Situation 13& RPCA        & -                              & F     & Mode 2   \\
            Situation 14& RPCA        & -                              & F     & Mode 3   \\
            Situation 15& IMPPCA    & Modes 1, 2, 3           & H     & Mode 1   \\
            Situation 16& IMPPCA    &  -                             & H     & Mode 2 \\
            Situation 17& IMPPCA    & -                              & H    & Mode 3 \\
  			\hline
  		\end{tabular}
  	\end{center}
  \end{table}

The monitoring results of Fault 4 and Fault 5 are summarized in Table \ref{Table2-comparative}. Take Fault 4 as an instance to explain the results detailedly. SPCA can detect the fault in Mode 1 accurately, but the FAR is $5.62\%$. When a new mode 2 arrives, the  model is updated based on the model A and the newly collected data. Thus, the monitoring model B is able to monitor the two successive modes simultaneously, and the FDRs are higher than $99\%$. Besides, the FAR of Situation 3 is lower than that of Situation 1, which indicates that information of Mode 2 is beneficial to reduce the false alarms of Mode 1.
The monitoring model C  enables to monitor two modes. But the FAR of Mode 1 is $6.5\%$ and  higher than that of SPCA-SI. When the new mode 3 appears, the proposed SPCA-SI trains the model D based on the model B and the current data collected. It enables to monitor the three modes simultaneously and the FDRs approach to $100\%$. Besides, the FARs are the lowest among all situations.
It is revealed that  SPCA-SI can preserve partial significant information of trained modes, which is advantageous to monitor other similar  modes.
RPCA fails to detect the fault in three modes and the FARs approximate to $100\%$. Although IMPPCA is capable of monitoring modes 1 and 2, the FAR for Mode 3 is $41.12\%$. IMPPCA fails to deliver desired expert level monitoring performance. 

\begin{figure*}[!tbp]
	\centering
	%
	\hspace{-1mm}
	\subfigure{\label{fault4-2}}\addtocounter{subfigure}{-2}
	\subfigure
	{\subfigure[Situation 2]{\includegraphics[width=0.235\textwidth]{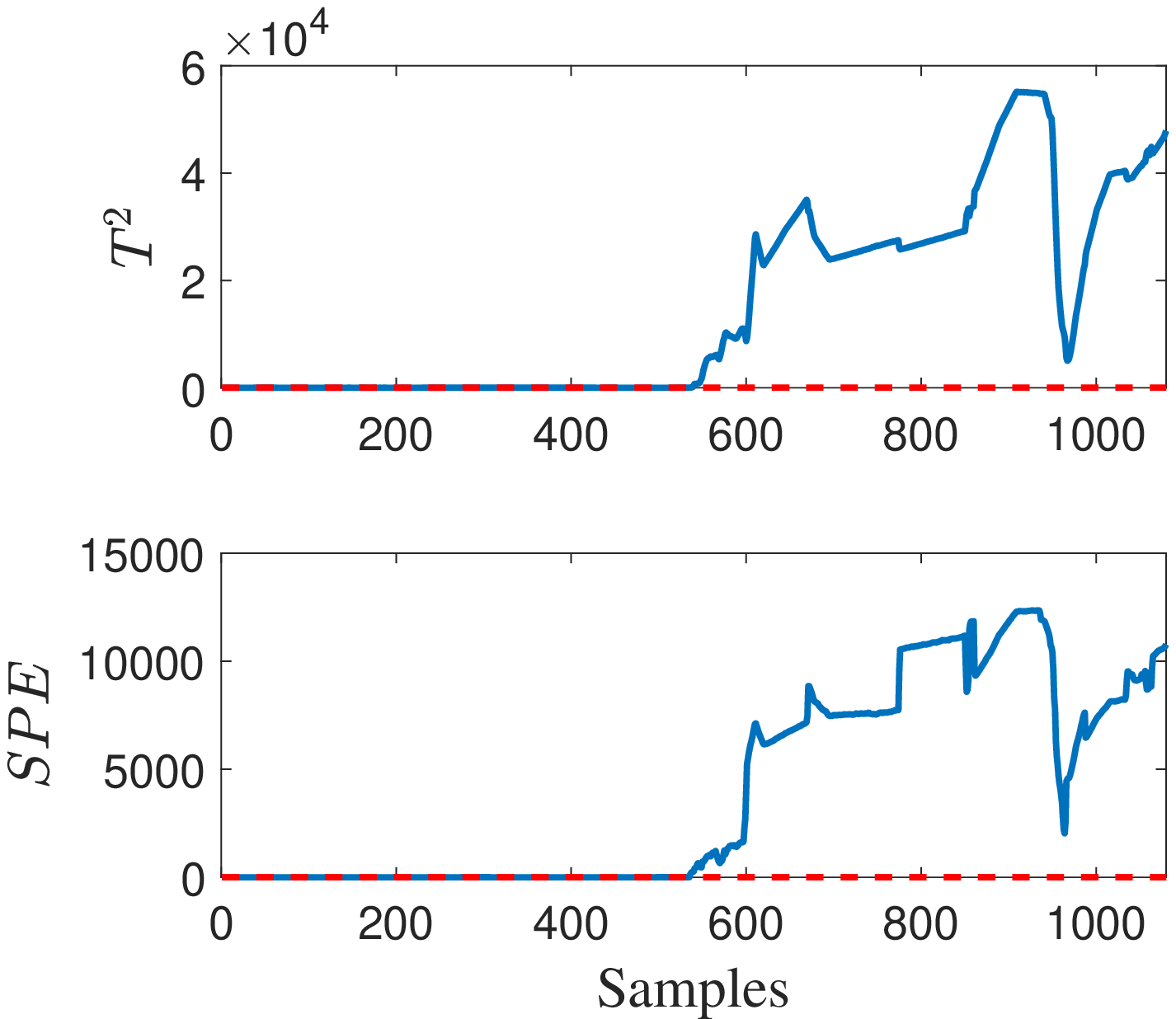}}}
     \subfigure{\label{fault4-3}}\addtocounter{subfigure}{-2}
       \hspace{-1mm}
	\subfigure
	{\subfigure[Situation 3]{\includegraphics[width=0.235\textwidth]{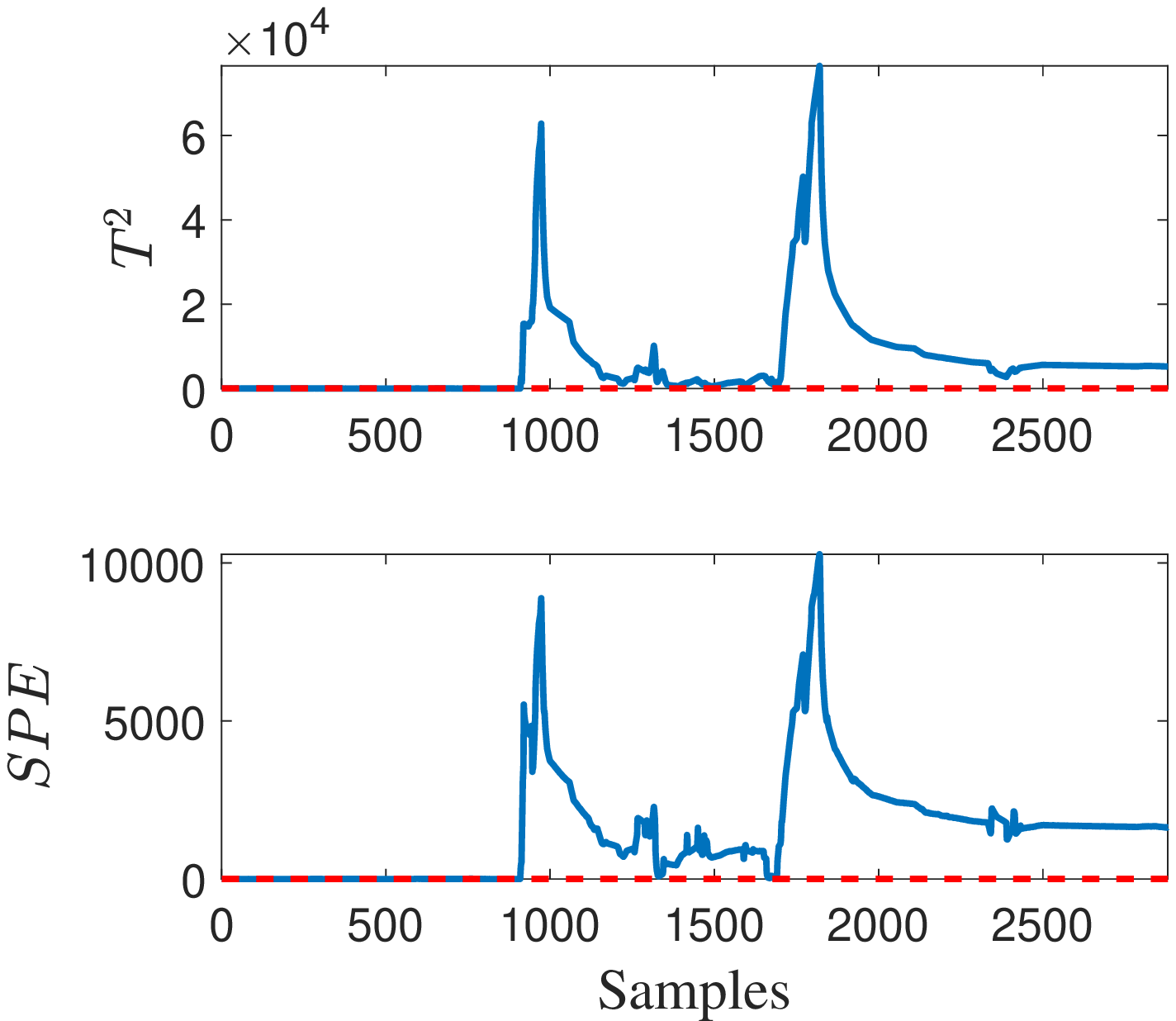}}}
		\hspace{-1mm}
%
    \subfigure{\label{fault4-6}}\addtocounter{subfigure}{-2}
    	\subfigure
    	{\subfigure[Situation 6]{\includegraphics[width=0.235\textwidth]{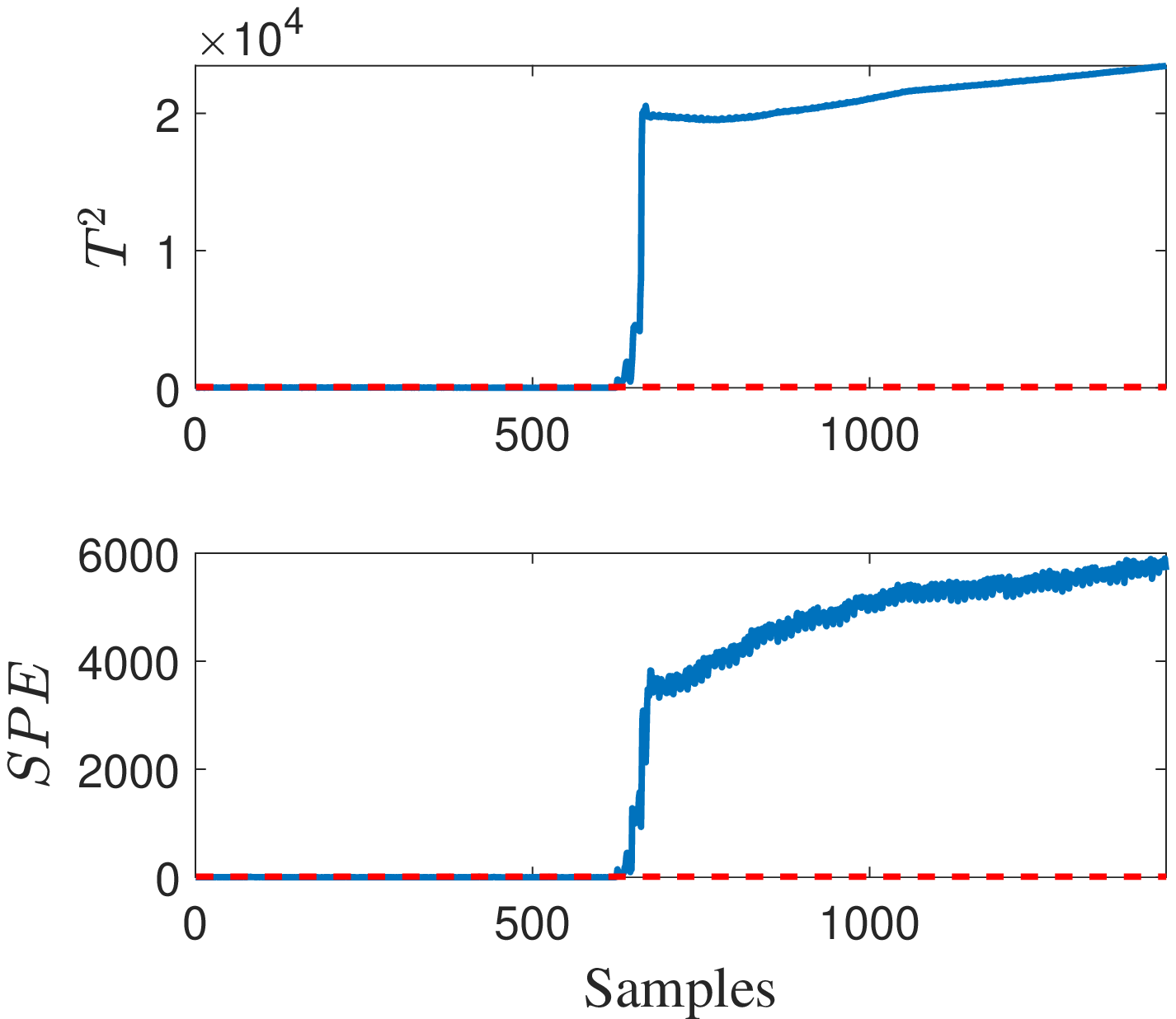}}}
    \subfigure{\label{fault4-7}}\addtocounter{subfigure}{-2}
     	\subfigure
    	{\subfigure[Situation 7]{\includegraphics[width=0.235\textwidth]{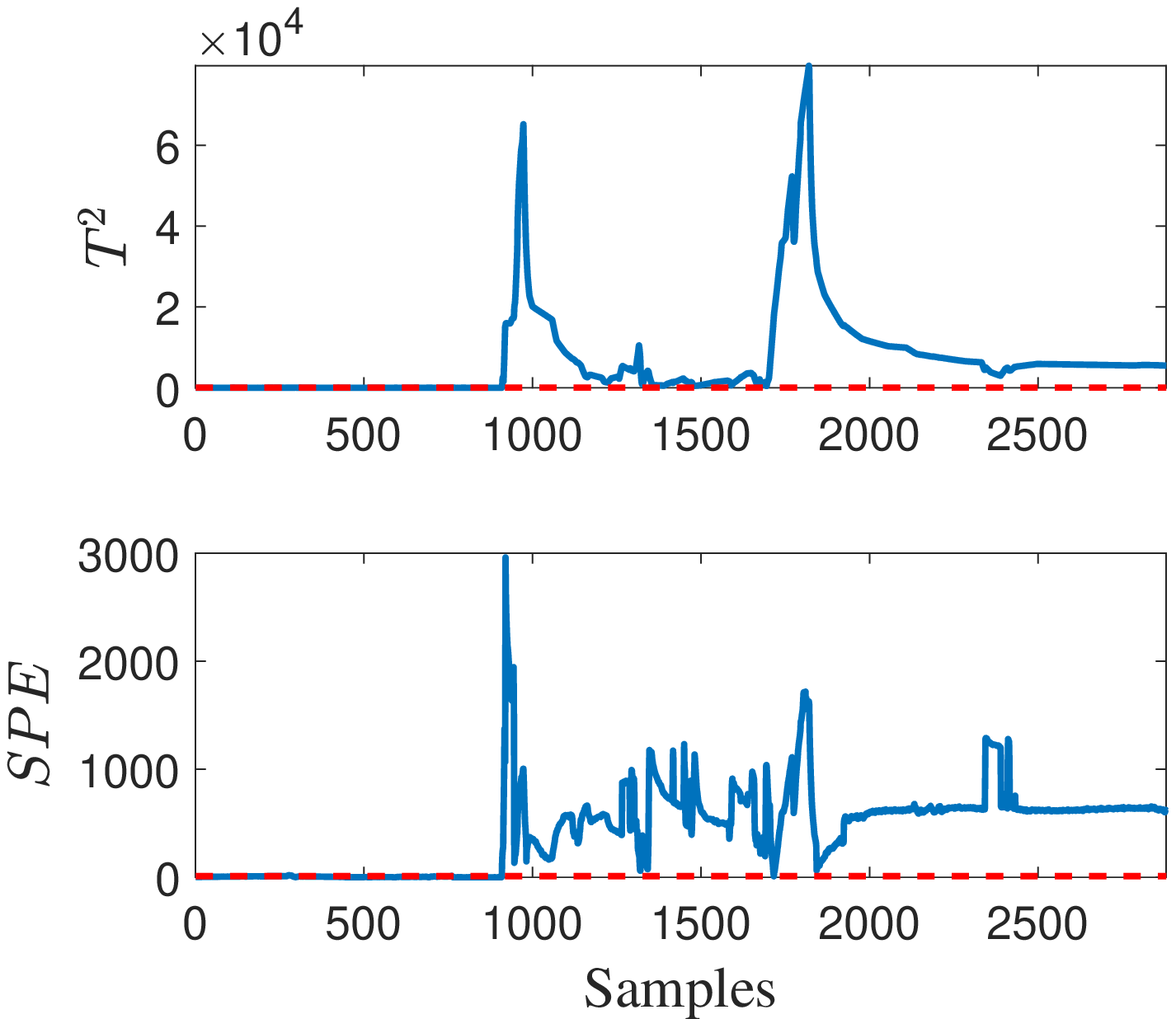}}}
    \subfigure{\label{fault4-8}}\addtocounter{subfigure}{-2}
     	\subfigure
    	{\subfigure[Situation 8]{\includegraphics[width=0.235\textwidth]{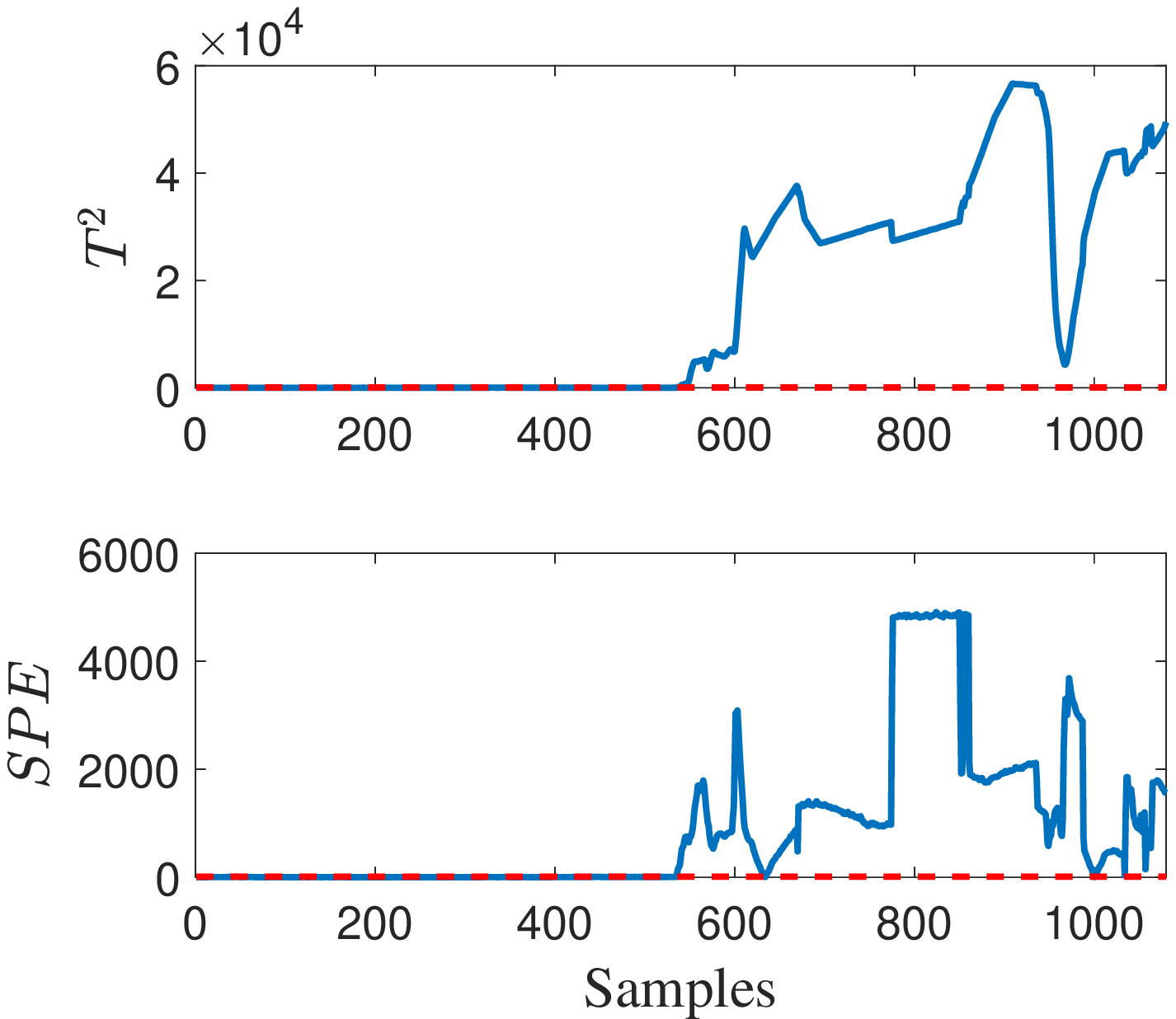}}}
    \subfigure{\label{fault4-11}}\addtocounter{subfigure}{-2}
     	\subfigure
    	{\subfigure[Situation 11]{\includegraphics[width=0.230\textwidth]{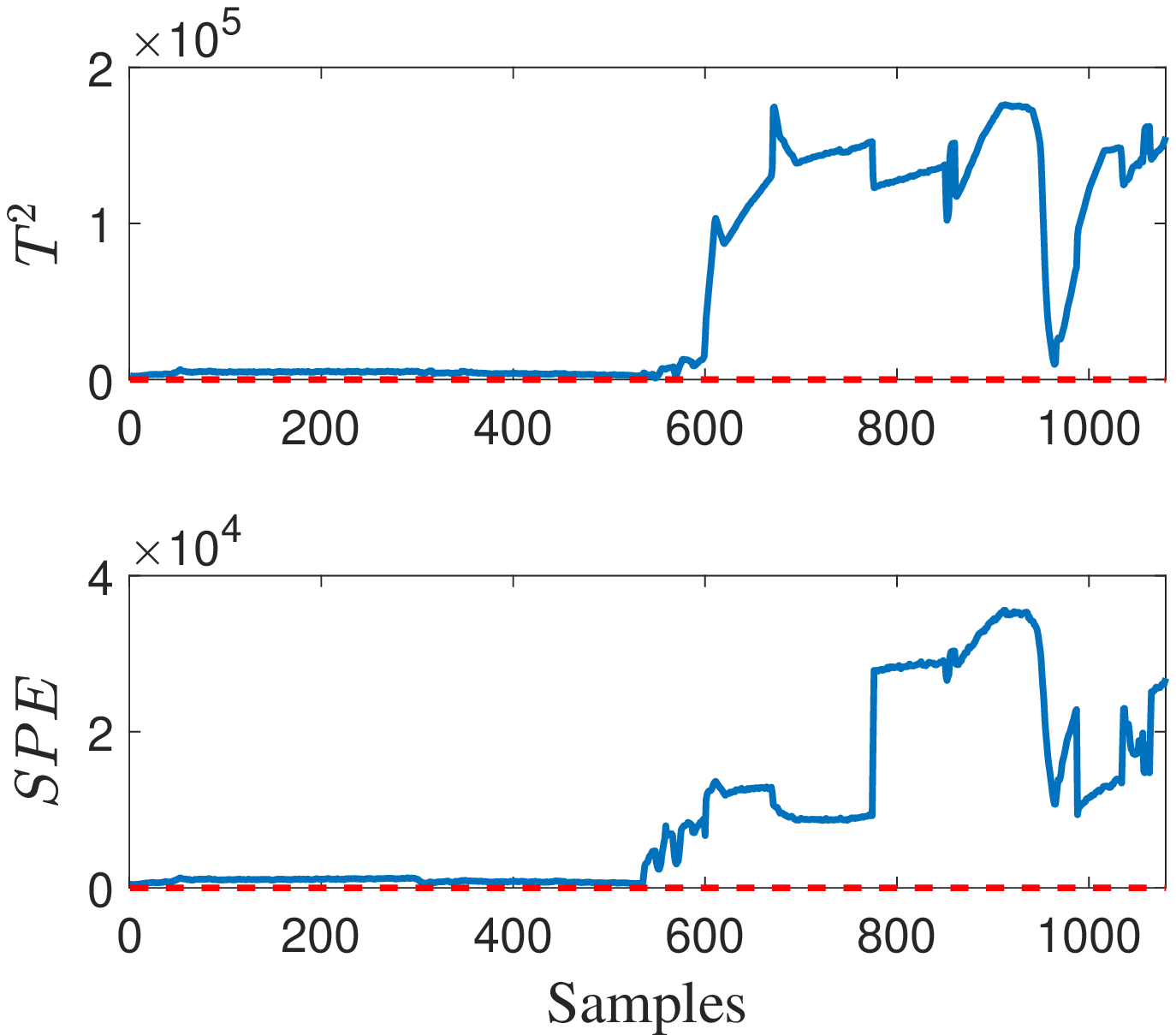}}}
\subfigure{\label{fault4-14}}\addtocounter{subfigure}{-2}
\subfigure
    	{\subfigure[Situation 14]{\includegraphics[width=0.235\textwidth]{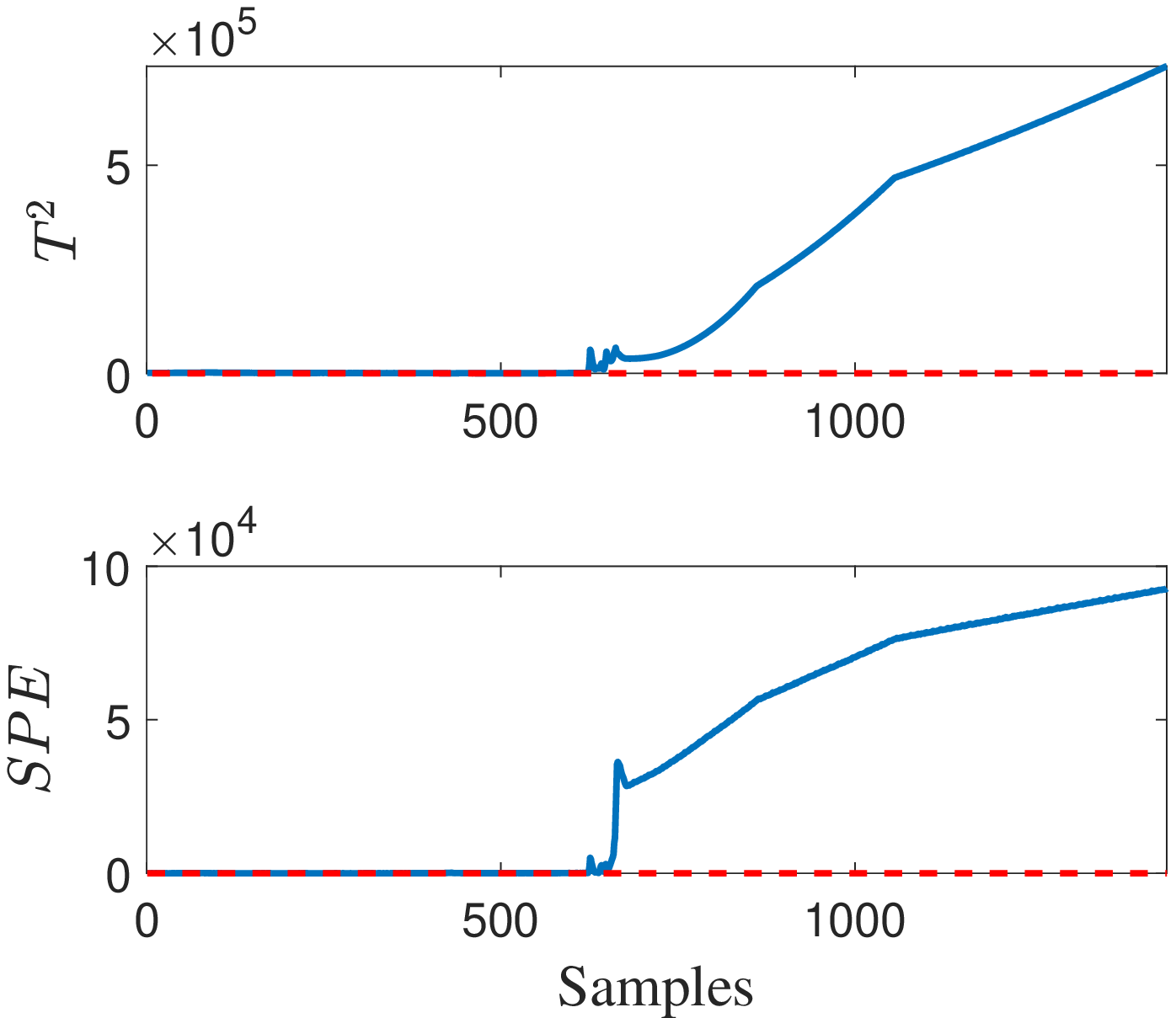}}}
\subfigure{\label{fault4-17}}\addtocounter{subfigure}{-2}
\subfigure
    	{\subfigure[Situation 17]{\includegraphics[width=0.235\textwidth]{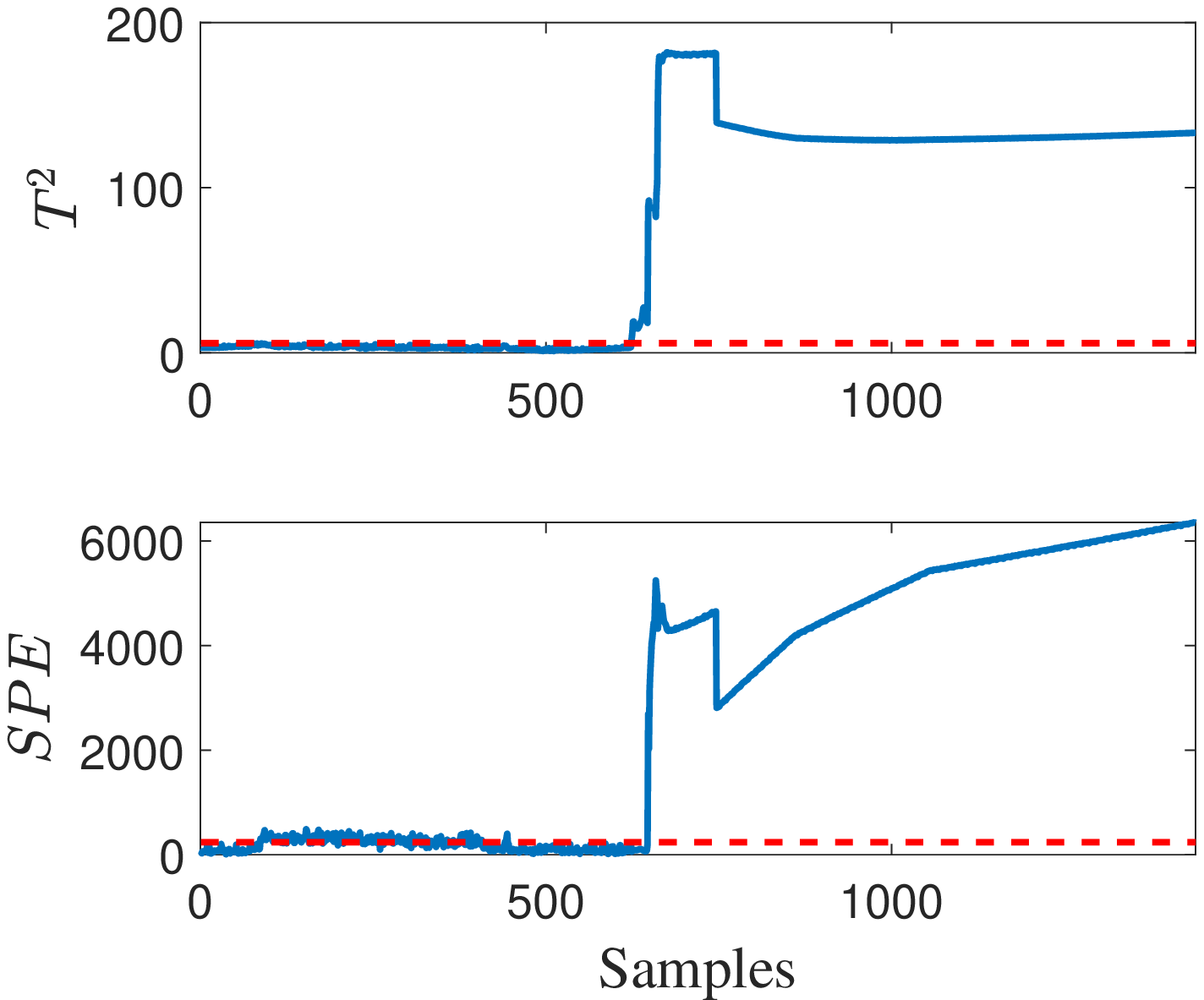}}}
	\centering
	\caption{Monitoring charts of Fault 4} \label{case4}
\end{figure*}

Owing to the paper length limitation, we just select 8 representative monitoring charts of Fault 4, as exhibited in Fig. \ref{case4}. The simulation results of SPCA-SI are mainly listed. Two charts of  RPCA and IMPPCA  are selected as comparison. The analysis of Fault 4 also applies to Fault 5. Note that the FAR of  IMPPCA  for Mode 2 is $51.28\%$.

According to Table \ref{Table1-comparative2}, SPCA is vulnerable to catastrophic forgetting issue and fails to monitor multiple modes based on the same model. SPCA-SI provides self-learning ability and the model is updated when a new mode arrives, which enables it to monitor multiple modes accurately. Moreover, the learned knowledge of previous modes is preserved continually and the model-sensitive parameters are fewer than PCA.
Similar to numerical case, RPCA is incapable of separating normal modes and faults. Besides, IMPPCA is unable to monitor three modes accurately as the FARs are more than $20\%$. When a new mode appears, we need to store data and retrain the model from scratch, which costs considerable  resources and energy. In conclusion, SPCA-SI  outperforms other comparative methods in consideration of detection accuracy and demanding resources in the long term.

\begin{table}[!tbp]
  	\begin{center}
  		\caption{FDR ($\%$) and FAR ($\%$) for the practical case}\label{Table2-comparative}
  		\footnotesize
  		\begin{tabular}{c c c c c }
  			\hline
       Fault type  &  \multicolumn{2}{c}{Fault 4} &  \multicolumn{2}{c}{Fault 5}  \\
        \cmidrule(r){1-1}  \cmidrule(r){2-3}  \cmidrule(r){4-5}
        Indexes      &   {FDR}    & {FAR} &   {FDR}    & {FAR}\\
        \hline
        Situation 1  & 99.95   &5.62     & 100        &  0      \\
        Situation 2  & 99.45   &0        &100         & 5.98    \\
        Situation 3  & 99.95   & 4.07    & 100        & 0       \\
        Situation 4  & 99.45   & 0       & 100        & 15.38   \\
        Situation 5  &99.95    & 6.5     & 100        & 0       \\
        Situation 6  &100      & 0.32    & 93.49      & 0       \\
        Situation 7  &99.95    & 1.54    & 100        & 0       \\
        Situation 8  &99.45    & 0       & 100        & 13.96  \\
        Situation 9  &100      & 0.48    & 92.75      & 0       \\
        Situation 10 &99.95    & 75.77   & 100        & 0      \\
        Situation 11 & 100     & 100     & 100        & 94.87    \\
        Situation 12 & 100     & 99.45   & 100        & 100  \\
        Situation 13 & 100     & 100     & 100        & 100  \\
        Situation 14 & 100     & 100     & 100        & 100  \\
        Situation 15 & 100     & 6.61    & 100        & 3.23  \\
        Situation 16 & 100     & 6.95    & 100        & 51.28  \\
        Situation 17 & 100     & 41.12   & 98.96      & 0   \\
    	\hline
  		\end{tabular}
  	\end{center}
  \end{table}

\section{Conclusion}\label{sec:5}


This paper presented a novel SPCA-SI method with self-learning ability for monitoring successive modes. The importance measure of variables is evaluated by SI along the learning trajectory. The acquired knowledge of previous modes is accumulated and the model is updated when new data are available, thus delivering excellent performance for successive modes.  The optimization issue is settled by APG and the learning rate is adaptively determined  to accelerate convergence. 
Besides, the influence of parameters is discussed and different methods can be converted by specific parameter setting. SPCA-SI furnishes excellent model interpretability, as the critical parameters are sparse. Moreover, a novel $T^2$ statistic is presented, where the significant information of previous modes is consolidated further.  
The effectiveness of the proposed method has been illustrated by a numerical case and a practical industrial system.

This proposed  method requires the similarity among different modes and prior information about mode switching time. In future, we'll investigate the numerous and diverse modes, with the mode switching time identified automatically.

\bibliographystyle{IEEEtran}
\bibliography{my_references}
\end{document}